\documentclass[10pt,twocolumn,letterpaper]{article}

\usepackage[pagenumbers]{cvpr}              %
\usepackage[dvipsnames]{xcolor}

\definecolor{cvprblue}{rgb}{0.21,0.49,0.74}
\usepackage[pagebackref,breaklinks,colorlinks,citecolor=cvprblue]{hyperref}

\newcommand\blfootnote[1]{%
  \begingroup
  \renewcommand\thefootnote{}\footnote{#1}%
  \addtocounter{footnote}{-1}%
  \endgroup
}
\newcommand{\benchmarkname}{UniQA-3D}

\title{Towards Foundation Models for 3D Vision: How Close Are We?}

\author{Yiming Zuo*, Karhan Kayan*, Maggie Wang, Kevin Jeon, Jia Deng, Thomas L. Griffiths\\
Princeton University\\
{\tt\small \{zuoym,karhan,maggiewang,kevinjeon,jiadeng,tomg\}@princeton.edu}}

\begin{document}

\maketitle

\begin{abstract}
  Building a foundation model for 3D vision is a complex challenge that remains unsolved. Towards that goal, it is important to understand the 3D reasoning capabilities of current models as well as identify the gaps between these models and humans. Therefore, we construct a new 3D visual understanding benchmark named \benchmarkname. \benchmarkname\ covers fundamental 3D vision tasks in the Visual Question Answering (VQA) format. We evaluate state-of-the-art Vision-Language Models (VLMs), specialized models, and human subjects on it. Our results show that VLMs generally perform poorly, while the specialized models are accurate but not robust, failing under geometric perturbations. In contrast, human vision continues to be the most reliable 3D visual system. We further demonstrate that neural networks align more closely with human 3D vision mechanisms compared to classical computer vision methods, and Transformer-based networks such as ViT~\cite{dosovitskiy_image_2021} align more closely with human 3D vision mechanisms than CNNs. We hope our study will benefit the future development of foundation models for 3D vision. Code is available at \url{https://github.com/princeton-vl/UniQA-3D}.
\end{abstract}

\blfootnote{*These authors contributed equally (random order).}

\vspace{-8mm}

\section{Introduction}
In recent years, impressive improvements in model accuracy and generalization ability on 2D vision tasks have been achieved with the introduction of \textit{foundation models}. Vision-Language Models (VLMs) such as GPT4~\cite{openai2023gpt} and LLaVA~\cite{li2023llava} can solve a wide range of visual understanding tasks, including Visual Question Answering (VQA) and image captioning on diverse datasets~\cite{liu2024LLavabenchmark}. To thoroughly evaluate the performance of these foundation models, several benchmarks have been proposed~\cite{lu2024wildvision,bitton2023visit-benchmark,ging2024open-benchmar,liu2023mmbench,li2023seedbench,li2024seedbench2,fu2024blink}. Among them, a few benchmarks compare the VLM performance against humans~\cite{fu2024blink,zhang2016vqa,goyal2017making}. Such comparison is crucial for understanding the robustness of the models and their alignment with human judgments in related tasks.

In comparison, foundation models and benchmarks are largely absent for 3D vision. On one hand, researchers often focus on training \textit{specialized models} that can solve only a single task, such as depth estimation~\cite{marigold,Ranftl2022midas} or optical flow~\cite{teed_raft_2020}. On the other hand, each task has its own evaluation metrics and there is no benchmark for comparing a single model across different tasks, since the output space of each task is vastly different (\textit{e.g.}, pixel-wise for depth estimation versus $SE(3)$ for camera pose estimation). The pixel-wise dense output required by many tasks also poses challenges in evaluating human performance, making it difficult to study and understand the differences and similarities between models and the human visual system.

To develop a foundation model for 3D vision, we must first understand the 3D vision capabilities of the existing models. Therefore, in this paper, we thoroughly evaluate the 3D understanding capabilities of a wide variety of models and focus on answering the following core questions: 

\begin{itemize}[leftmargin=*]
  \item Do 2D VLMs show emergent 3D understanding capability allowing them to solve 3D tasks?

  \item What are the accuracy and robustness of specialized models trained on each task?
  
  \item Do humans remain the most accurate and robust 3D vision system? How do the error patterns of VLMs and specialized models compare to those of humans?

\end{itemize}

To answer the above questions, we construct a new 3D visual understanding benchmark \benchmarkname\ (\textbf{UNI}fied Visual-\textbf{Q}uestion-\textbf{A}nswering for \textbf{3D} Vision), which covers fundamental 3D vision tasks including depth estimation, spatial reasoning, camera pose estimation, and keypoint matching. While our benchmark is based on existing datasets with accurate ground truth, its key feature is a \textit{unified output space} across tasks. We formulate all the questions in VQA format so that they can be easily answered by both VLMs and humans, enabling fair comparisons. For example, our depth estimation benchmark asks for binary depth relationships between two pixels rather than a dense depth map. Furthermore, we construct challenging cases with \textit{geometric perturbation}, such as flipping or rotating the images, to make them diverge from the gravity-aligned views on which the models were trained. Those geometric perturbations are common in applications such as robotics, making it important to test the robustness of models in such scenarios. Compared to existing benchmarks, \benchmarkname\ is the first to focus entirely on 3D, as shown in~\cref{tab:benchmark_comparison}.

We conduct extensive experiments on \benchmarkname\ by evaluating a variety of models (Sec.~\ref{sec:experiments}). We test state-of-the-art VLMs including GPT4-Turbo, GPT4-Omni, and Gemini-1.5. We focus on closed-source VLMs because they have the best performance on the 2D leaderboards~\cite{bitton2023visit-benchmark,liu2023mmbench}. We also test the performance of state-of-the-art specialized models on each task. Finally, we record human performance using Amazon Mechanical Turk (MTurk).

Here we present our key findings and our answers to the above questions:

\textbf{Can 2D VLMs solve 3D tasks well?}
No. While VLMs achieve impressive performance on existing VQA benchmarks focusing on 2D, we find that they have \textit{poor} 3D understanding abilities. None of the existing  VLMs achieves human-level performance or performs on par with the specialized model. Surprisingly, on some tasks (\textit{e.g.,} depth estimation), the VLMs perform only marginally better than random guessing, and on geometrically perturbed images, they perform even worse than random guessing.

\textbf{Are specialized models accurate and robust?} They are accurate but not robust. The specialized models have high accuracy in general, and interestingly we find that they are even better than humans on some tasks (\textit{e.g}., depth estimation and spatial reasoning) in the zero-shot setting. However, they are not as robust as humans against geometric perturbations. For example, on depth estimation, the accuracy of MiDaS~\cite{Ranftl2022midas} drops significantly (from 90.4\% to 73.9\%) on the upside-down images, while human accuracy remains the same (about 83\%). Our finding suggests that the specialized models are still vulnerable, despite being trained on a large collection of diverse data~\cite{Ranftl2022midas,lindenberger_LightGlue_2023}.

\textbf{Are humans the most accurate and robust 3D visual system? How do the error patterns of models compare to those of humans?} Yes, humans remain the most accurate and robust 3D visual system. Compared to humans, the error patterns of different models vary significantly according to the model types and architectures. Here are our findings: 1) specialized models have better alignment with humans compared to VLMs according to Cohen's $\kappa$~\cite{mchugh2012interrater}; 2) the error patterns of neural networks are more similar to humans than hand-crafted algorithms such as SIFT~\cite{sift} in the context of keypoint detection and correspondence matching; and 3) in terms of model architecture, the error patterns of the Transformer-based ones (\textit{e.g.}, ViT~\cite{dosovitskiy_image_2021}) are more similar to humans compared to the CNN-based ones. While Tuli \textit{et al.}~\cite{Tuli2021AreCN} found that Transformers are more similar to humans on 2D classification tasks than CNNs, the generalization of this finding to 3D is valuable and non-trivial. This may be because the mechanisms for processing 2D and 3D in the human brain are different: 2D vision is handled by the ventral stream in the brain while 3D vision is performed by the dorsal stream~\cite{dicarlo_how_2012, welchman_human_2016}.

Our main contributions are as follows:
\begin{itemize}[leftmargin=*]
  \item We propose a new benchmark \benchmarkname\ with a unified output space for fair comparison of the 3D understanding capability of different models and humans. 
  
  \item We evaluate the performance of state-of-the-art VLMs, specialized models, and humans on our benchmark, and we compare the error patterns of different models against humans under multiple criteria.
\end{itemize}
 We hope our results will benefit the future development of foundation models for 3D vision by providing insights for improving their robustness and ability to generalize.

\begin{table}[t]
\setlength{\tabcolsep}{1.0mm}
\centering
        \caption{\benchmarkname\ has comparable size (in terms of the number of images) to modern benchmarks for testing VLMs, such as MMBench~\cite{liu2023mmbench}. Moreover, \benchmarkname\ provides human accuracy and contains challenging samples with geometric perturbation.}
        \label{tab:benchmark_comparison}
\begin{tabular}{lcccc}
\hline
Benchmarks & Modality & \#Images & \begin{tabular}{@{}c@{}}{Human} \\ {Acc.} \end{tabular} & \begin{tabular}{@{}c@{}}{Geom.} \\ {Perturb} \end{tabular} \\
\hline
\hline
VQA~\cite{antol2015vqa} & 2D & 250,000 & Yes & No \\
MSCOCO~\cite{lin2014mscoco} & 2D & 328,000 & No & No \\
MMVet~\cite{yu2023mmvet} & 2D & 200 & No & No \\
MMBench~\cite{liu2023mmbench} & 2D & 2,590 & No & No \\
SeedBench~\cite{li2023seedbench} & 2D & 19,242 & No & No \\
VisIT~\cite{bitton2023visit-benchmark} & 2D & 592 & No & No \\ 
BLINK~\cite{fu2024blink} & 2D+3D & 7,358 & Yes & No \\
\hline
\textbf{\benchmarkname\ (Ours)} & 3D & 2,450 & Yes & Yes \\
\hline
\end{tabular}
\vspace{-5mm}
\end{table}

\section{Related Work}

\subsection{VLMs and Benchmarks} 

Vision Language Models (VLMs) are typically trained on large-scale datasets of text-image pairs and achieve impressive performance and generalization on image understanding~\cite{lu2024wildvision,bitton2023visit-benchmark}. There are both open-sourced VLMs such as LLava~\cite{li2023llava}, CogVLM~\cite{wang2023cogvlm}, and MiniCPM~\cite{hu2024minicpm}, and  closed-sourced VLMs that one can only access through paid API, such as GPT4~\cite{openai2023gpt}, Claude~\cite{bai2022claude}, and Gemini~\cite{team2023gemini}. In our evaluation, we focus mainly on closed-sourced models because they have better performance in general~\cite{bitton2023visit-benchmark}.

Several works combine VLMs with 3D understanding. For instance, 3D-LLM~\cite{3dllm} and 3D-VisTA~\cite{3dvista} train an LLM that can take 3D point clouds as input and complete high-level tasks (\textit{e.g.}, navigation) by leveraging paired 3D-language datasets. In contrast, our paper focuses on benchmarking existing models instead of training. Additionally, we use 2D images as input modality and focus more on low-level 3D vision tasks. Banani \textit{et al.}~\cite{el2024probing} trains linear probes on the features of large models to solve 3D tasks such as depth estimation and matching. Compared to them, evaluating on our \benchmarkname\ requires no training and can be applied to closed-sourced VLMs and humans.

Various benchmarks have been proposed to test the scene understanding and visual question answering capability of 2D VLMs~\cite{lu2024wildvision,bitton2023visit-benchmark,ging2024open-benchmar,fu2024blink,liu2023mmbench,li2023seedbench,li2024seedbench2,fu2024blink}. Datasets from a decade ago such as MSCOCO~\cite{lin2014mscoco} and VQA~\cite{antol2015vqa} are large in scale but lack robust and fine-grained evaluation. To resolve this, MMBench~\cite{liu2023mmbench} proposes an objective evaluation scheme with fine-grained classes of questions. VisIT~\cite{bitton2023visit-benchmark} covers 70 families of instructions and proposes an automatic LLM-based evaluation aligned with human preferences. WildVision~\cite{lu2024wildvision} generates open-ended questions automatically by leveraging classification datasets. All of these datasets focus on testing the 2D capability of VLMs, whereas \benchmarkname\ focuses entirely on 3D. 

BLINK~\cite{fu2024blink} evaluates on a diverse set of tasks. Some of the tasks are in 3D (\textit{e.g.,} relative depth), while others are in 2D (\textit{e.g.,} IQ test). In contrast to BLINK, our paper focuses entirely on 3D, covering more 3D tasks such as keypoint detection and including more challenging samples with geometric perturbations. Moreover, BLINK evaluates human performance on only 2 subjects (the coauthors), whereas we evaluate human performance with 162 subjects in depth estimation, 109 subjects in camera pose estimation, 449 subjects in VQA, and 143 subjects in keypoint matching. The large sample size of our \benchmarkname\ helps us discover statistically significant differences between models and humans. 

\subsection{Cognitive Science with Neural Networks}
Cognitive scientists have extensively studied the relationship between computer vision models and the human visual system in the context of 2D vision, particularly object recognition~\cite{yamins_performance-optimized_2014, khaligh-razavi_deep_2014, rajalingham_large-scale_2018, cichy_comparison_2016, schrimpf_brain-score_2020}. However, a similar comparison is missing for 3D vision. Most research on the human 3D visual system has focused on Marr's implementation level~\cite{marr_vision_2010}, such as the discovery of grid cells, place cells, and head-direction cells explaining which parts of the brain activate for visual localization~\cite{barry_neural_2014}, or the stereogram and fMRI based works explaining which parts of the brain activate for depth perception~\cite{welchman_human_2016}. However, these works are unlikely to provide direct insight into constructing 3D foundation models. 

Our paper aims to provide such insight by quantitatively comparing the performance and error patterns of 3D vision models to human subjects. In the context of object recognition, Geirhos et al.~\cite{geirhos_beyond_2020} introduced Cohen's $\kappa$ analysis in order to compare neural network error patterns to humans, and Tuli et al.~\cite{Tuli2021AreCN} showed that human error patterns are more similar to ViTs than CNNs. However, the brain understands 2D and 3D through two different streams~\cite{dicarlo_how_2012, welchman_human_2016}, suggesting that these findings might not be simply extrapolated to 3D vision. Our findings show that human error patterns are, indeed, more similar to Transformers than CNNs in the case of depth perception. This is surprising considering the biological plausibility of convolution~\cite{Fukushima1980NeocognitronAS, yamins_performance-optimized_2014, richards_deep_2019}, particularly for depth perception~\cite{ohzawa_stereoscopic_1990}. Perhaps more surprisingly, we find that neither Transformers, nor CNNs perform camera pose estimation similar to humans, with the highest Cohen's $\kappa$ being 0.16. \looseness=-1

\subsection{Specialized 3D Vision Models} 

\noindent\textbf{Monocular Depth Estimation.} MiDaS~\cite{Ranftl2022midas} is the first to explore large-scale training on a mixture of datasets, followed by more recent works~\cite{yang2024depth,marigold}. Since absolute depth is challenging to predict, DIW~\cite{chen2016single} explores using humans to annotate the relative depth between two pixels. Our relative depth prediction task is inspired by DIW, but our relative depth ground-truth labels are generated from Lidar sensors rather than from human annotators, making them more reliable. We use MiDaS~\cite{Ranftl2022midas} in this study because it provides both CNN-based and ViT-based variants.

\noindent\textbf{Visual Question Answering (VQA).}  Tranformer-based VQA methods can be classified into single-stream~\cite{chen2019uniter,li2019visualbert,zhang2021vinvl,li2020oscar} and two-streams~\cite{lu2019vilbert,su2019vlbert,lu202012-in-1,tan2019lxmert}, depending on how the images and questions are processed. MDETR~\cite{kamath2021mdetr} explicitly detects the objects and fuses the vision and language stream with a Transformer. We use MDETR as the specialized model for its state-of-the-art accuracy. 

Many datasets have been proposed as VQA benchmarks: some focus on semantics~\cite{antol2015vqa,kafle2017tdiuc,shah2019kvqa,wang2017fvqa,ren2015cocovqa} while others focus on spatial relationship reasoning~\cite{johnson2017clevr,yi2019clevrer,hudson2019gqa}. We build our benchmark on top of the CLEVR~\cite{johnson2017clevr} dataset. 

\noindent\textbf{Camera Pose Estimation} aims to estimate the 6DoF pose (translation and rotation) of a camera either with respect to a global coordinate system or relative to another camera. Matching-based methods estimate the relative pose using matched keypoints~\cite{schoenberger2016sfm, mur-artal_orb-slam_2015, teed_deep_2023, teed_droid-slam_2022, sarlin_coarse_2019, 8237522, sarlin_coarse_2019, panek_meshloc_2022}, whereas pose regression methods~\cite{arandjelovic_netvlad_2016, kendall_posenet_2016, walch_image-based_2017, chen_dfnet_2022, laskar_camera_2017, shavit_learning_2021, chen_direct-posenet_2021} estimate the 6DoF pose directly based on the input images. Our experiments show that none of the models classify pose similar to humans with the highest Cohen's $\kappa$ score being 0.16. 

\noindent\textbf{Keypoint Detection} involves detecting salient keypoints from an image, typically followed by local feature extraction around those keypoints. Classical computer vision methods such as~\cite{sift, Harris1988ACC, Rosten2006MachineLF, Rublee2011ORBAE} extract keypoints and features using local information such as image gradients. In contrast, recent deep neural network-based methods such as~\cite{detone_superpoint_2018, barroso-laguna_keynet_2019, tyszkiewicz_disk_2020, dusmanu_d2-net_2019} train CNNs to detect keypoints and extract local features. Our results suggest that humans follow a keypoint detection strategy more similar to neural networks than classical methods. 

\noindent\textbf{Keypoint Matching} matches a pixel to the corresponding pixel in another view. Keypoint matching methods can be grouped into two classes: detector-free methods~\cite{sun_loftr_2021, teed_raft_2020, wang_matchformer_2022, chen_aspanformer_2022}, which perform dense matching and avoid the keypoint detection phase; and detector-based methods, which rely on a keypoint detector and local features extracted from those keypoints. Detector-based methods can be further divided into two subclasses: classical  methods~\cite{sift, Harris1988ACC, Rosten2006MachineLF, Rublee2011ORBAE, lowe_distinctive_2004}, which use k-Nearest Neighbors in the feature space; and deep learning-based matchers~\cite{lindenberger_LightGlue_2023, sarlin2020superglue, chen_learning_2021}, which train neural networks to match the extracted keypoints. Our experiments show that human keypoint matching is more similar to neural network-based methods than classical methods.\looseness=-1 

\section{The \benchmarkname\ Benchmark}
\label{sec:benchmark}

\begin{table}[t]
\setlength{\tabcolsep}{0.6mm}
\centering
        \caption{Statistics of \benchmarkname. Our benchmark has 4 sub-tasks and is built on top of existing datasets with high-quality annotations. $\dag$: We train custom models with the ResNet~\cite{he_deep_2015}, ViT~\cite{chen_learning_2021}, and Swin~\cite{liu_swin_2021} backbone. $\ddag$: For keypoint detection, we test SIFT~\cite{sift}, FAST~\cite{Rosten2006MachineLF}, and SuperPoint(SP)~\cite{detone_superpoint_2018}; for keypoint matching, we test ORB~\cite{Rublee2011ORBAE} and LightGlue~\cite{lindenberger_LightGlue_2023}.}
        \label{tab:benchmark_statistics}
    \vspace{-2mm}
\begin{tabular}{lccccc}
\hline
Task & \begin{tabular}{@{}c@{}}{Data} \\ {Source} \end{tabular} & \#Images &  \begin{tabular}{@{}c@{}}{Specialized} \\ {Models} \end{tabular} \\
\hline
\hline
 Relative Depth & KITTI~\cite{kitti} & 750 & MiDaS~\cite{Ranftl2022midas} \\
 Spatial Reasoning & CLEVR~\cite{johnson2017clevr} & 500 & MDETR~\cite{kamath2021mdetr} \\
 Camera Pose & DTU~\cite{jensen2014large} & 750 &  Custom$^\dag$ \\
\begin{tabular}{@{}c@{}}{Keypoint-} \\ {Matching} \end{tabular} & \begin{tabular}{@{}c@{}}{Megadepth} \\ {~\cite{li_megadepth_2018}} \end{tabular} & 450 & \begin{tabular}{@{}c@{}}{SIFT,FAST,SP} \\ {ORB,LighGlue$^\ddag$} \end{tabular} \\
\hline
\end{tabular}
\vspace{-5mm}
\end{table}

We develop a new benchmark \benchmarkname\ for 3D vision tasks based on existing public datasets (\cref{tab:benchmark_statistics}). The key feature of our benchmark is a \textit{unified output space} for all sub-tasks, \textit{i.e.,} we form all questions to be multiple-choice so that all models and humans can be easily and fairly compared. In addition to the specialized models, we collect the response from VLMs (GPT4-Turbo, GPT4-Omni, and Gemini-1.5) and humans.

We formally define the four tasks below. We use the term ``subject'' to refer to either a human or a model.

\subsection{Relative Depth Estimation}
\label{sec:benchmark/depth}

\textbf{Task Definition} The subject is provided a single image with two markers annotating two pixels in the image. The subject is asked to determine which pixel is closer to the camera. See \cref{fig:depth_input_example} for an example.

\noindent\textbf{Dataset} We use KITTI~\cite{Geiger2013IJRR} for the relative depth estimation analysis. KITTI is a real-world autonomous driving dataset, and we choose it for two reasons: 1) KITTI has rich and accurate annotations, including accurate depth collected by Lidar, as well as semantic segmentation labels. 2) KITTI images contain both natural and man-made objects, providing a high diversity of visual content. Following DIW~\cite{chen2016single}, we sample markers with a 50\% probability of being placed either randomly or symmetrically along a horizontal line. We collect 500 regular images and another 250 geometrically perturbed images by flipping them upside-down.

\subsection{Spatial Reasoning}
\textbf{Task Definition} The format of the questions follows the VQA task, but they specifically require the subject to reason about the spatial relationships among objects in a scene. See \cref{fig:clevr_example} for an example.

\noindent\textbf{Dataset} We use a subset of CLEVR~\cite{johnson2017clevr}, which is a synthetic dataset with complex spatial relationships. We use the ground-truth questions from CLEVR but only keep the questions that contain certain spatial keywords (\textit{e.g,} left, right, front, behind, top, bottom). Note our selected subset is significantly more challenging than the full CLEVR dataset: MDETR~\cite{kamath2021mdetr} achieves 99.7\% accuracy on the original CLEVR test set but only 74.4\% accuracy on our dataset. Our dataset contains 500 image-question pairs in total.

\subsection{Relative Camera Pose Estimation}
\textbf{Task Definition} We give the subject two views of the same scene with an object of focus. The subject is then asked to choose the most prominent motion of the camera, \textit{i.e.,} move left, right, down, or up.

\noindent\textbf{Dataset}
 We use DTU~\cite{jensen2014large}, which has accurate ground-truth for the poses of the camera mounted on a robotic arm. Out of the original 49 views, we randomly choose view pairs but ensure a prominent movement axis by setting a threshold on the ratio of the major movement axis to the minor movement axis. This ensures that the movement between the two views is clear and pronounced along a single axis, making the task easier to evaluate by both models and human subjects. We sample 500 image pairs and an additional 250 upside-down image pairs as geometric perturbations. 

\begin{figure*}[ht]
\vspace{-5mm}
    \centering
    \begin{subfigure}[b]{0.4\textwidth}
        \centering
    \includegraphics[width=\textwidth]{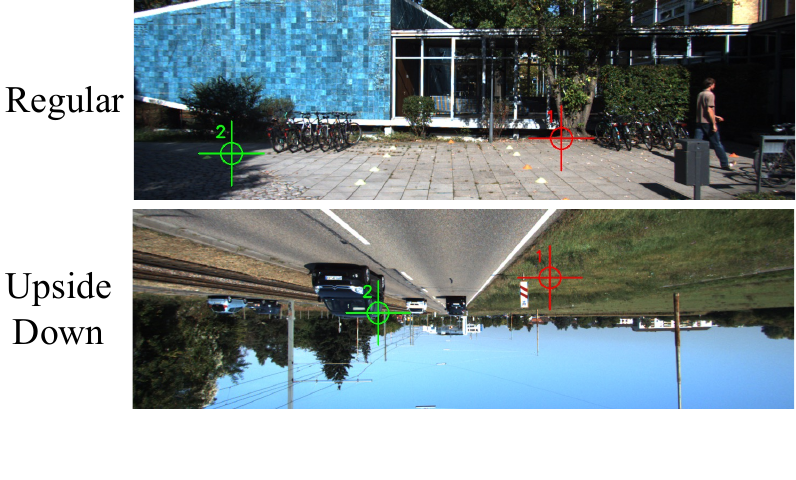}
        \caption{Each image has two markers and the subject decides which one is closer to the camera.}
        \label{fig:depth_input_example}
    \end{subfigure}
    \hfill
    \begin{subfigure}[b]{0.55\textwidth}
        \centering
    \includegraphics[width=\textwidth]{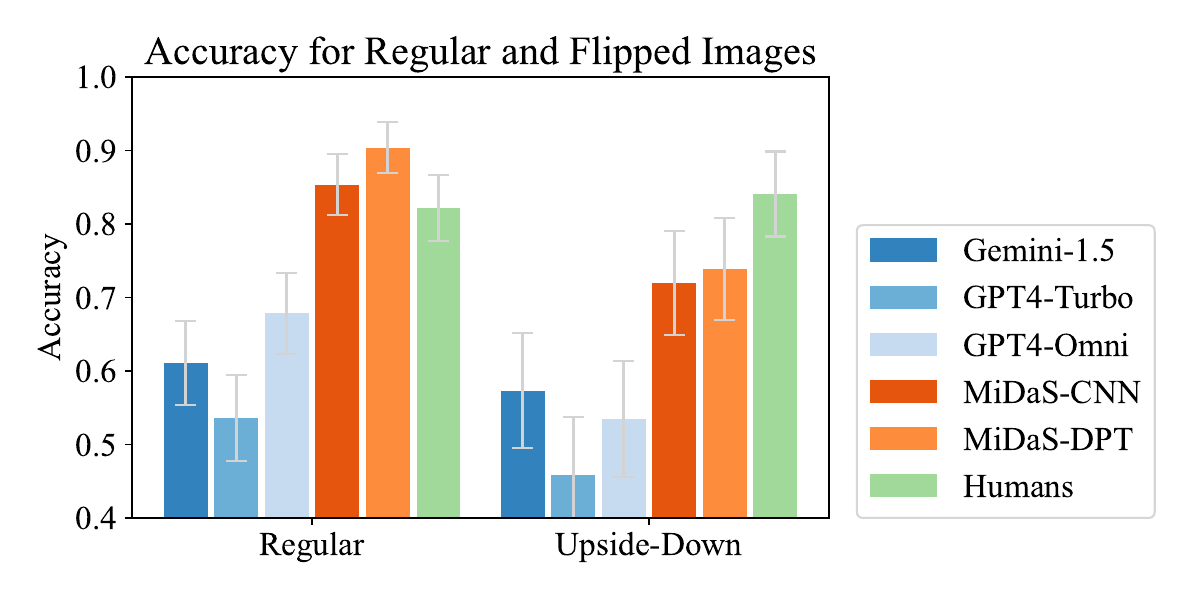}
        \caption{Accuracy on regular and upside-down images. MiDaS has comparable performance to humans while VLMs perform poorly.}
        \label{fig:depth_acc_regular_flipped}
    \end{subfigure}
    \hfill
    \vspace{1mm}
    \begin{subfigure}[b]{0.9\textwidth}
        \includegraphics[width=\linewidth]{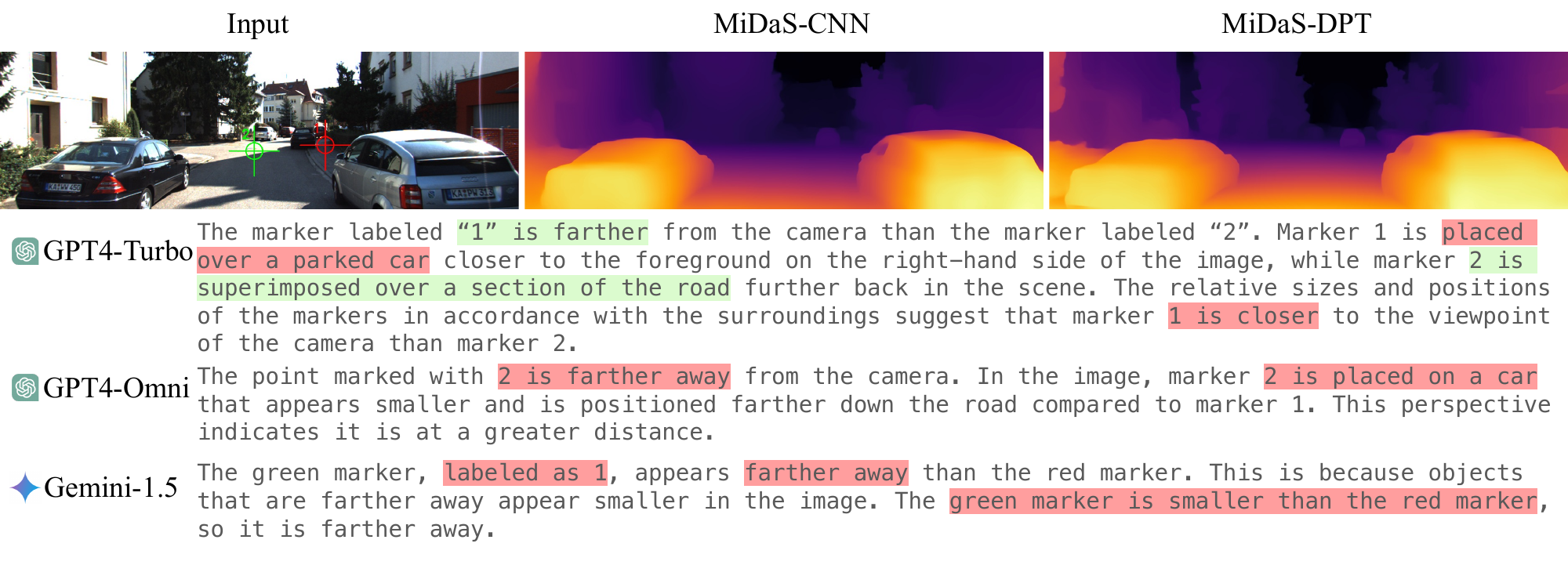}
        \caption{Output visualization. For VLM responses, we mark the correct statements green and the incorrect ones red. GPT4-Turbo and GPT4-Omni wrongly localize the markers, while Gemini-1.5 generates self-contradicting answers.}
        \label{fig:depth_acc_vis}
    \end{subfigure}
    \caption{(a) We sample images from the KITTI dataset and flip to create upside-down images. (b) Comparison of accuracy of different methods. MiDaS-DPT works the best in general, and both MiDaS models are slightly better than humans. All the VLMs perform poorly, with GPT4-Omni performing the best on regular inputs. (c) VLMs have multiple failure modes. See text for details.}
    \label{fig:depth_acc_robustness}
    \vspace{-4mm}
\end{figure*}

\subsection{Keypoint Matching}
\textbf{Task Definition}
We give the subject two views of a scene. We then ask the subject to choose five points in one image and match the corresponding points in the other image.

\noindent\textbf{Dataset}
We use the Megadepth-1500 dataset~\cite{sun_loftr_2021, li_megadepth_2018}. The scenes are historic locations with wide-baseline images. It provides accurate depth and camera pose ground-truth. For the keypoints chosen by the subject, we use the depth and camera pose to find the ground-truth correspondence in the other image, and we run additional forward-backward consistency checks to ensure co-visibility. We randomly sample 450 image pairs and ask the human subject to annotate 5 keypoint pairs on each, resulting in 2,250 keypoint pairs. 

\section{Experiment Results and Analysis}
\label{sec:experiments}
\subsection{Human Annotation Collection}
We collected human annotations using Amazon Mechanical Turk (MTurk) with IRB approval. Since there are bots/spammers on MTurk, it is important to filter them out to ensure high answer quality. We maintain strict quality control by: 1) requiring the HIT approval rate $\geq95$\% and that the number of HITs $\geq1000$; 2) when possible, structuring the multiple-choice questions to require clicking in designated locations, such as asking users to click inside a checkbox painted on the image. This allows us to filter out responses that do not follow the requested format; and 3) leveraging consensus scoring by assigning each HIT to 3 different users and only considering the result valid when all 3 users provide the same answer. 

\subsection{Relative Depth Estimation}

\textbf{Models Compared}  We use the same prompt for the VLMs as for the humans: ``There are two markers on the image. Which is farther away from the camera?" We also compare two variants of state-of-the-art specialized depth estimation model MiDaS~\cite{Ranftl2022midas}, \textit{i.e.}, MiDaS-CNN and MiDaS-DPT. We extract the relative depth relationship from the dense depth predictions. Note that MiDaS is not trained on KITTI. 

\begin{figure}[ht]
    \centering
    \begin{subfigure}[b]{0.49\linewidth}
        \centering
    \includegraphics[width=0.9\linewidth]{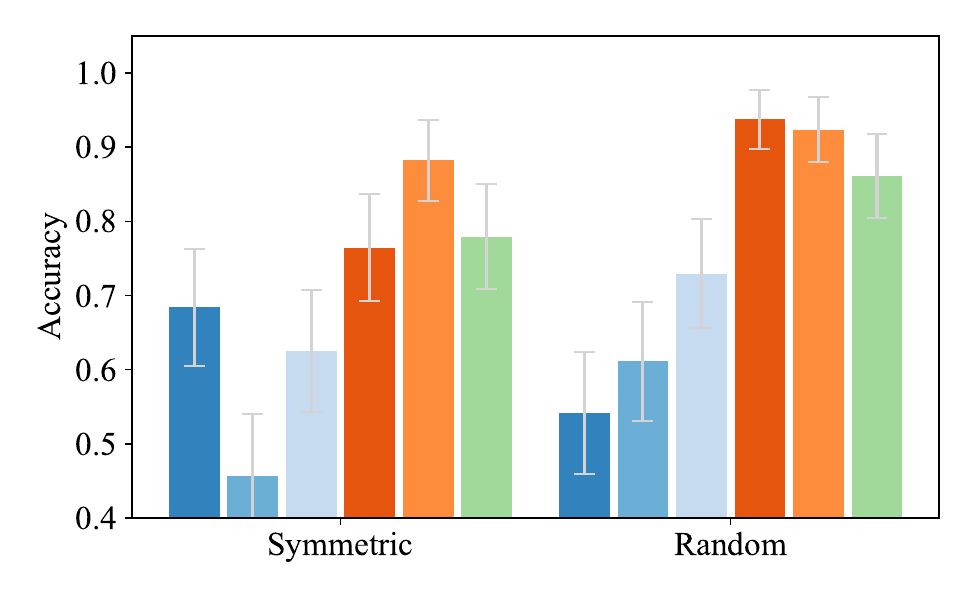}
        \caption{The model accuracy on symmetrically and randomly sampled pixel pairs.}
        \label{fig:depth_acc_symmrand}
    \end{subfigure}
    \hfill
    \begin{subfigure}[b]{0.49\linewidth}
        \centering
        \includegraphics[width=0.9\textwidth]{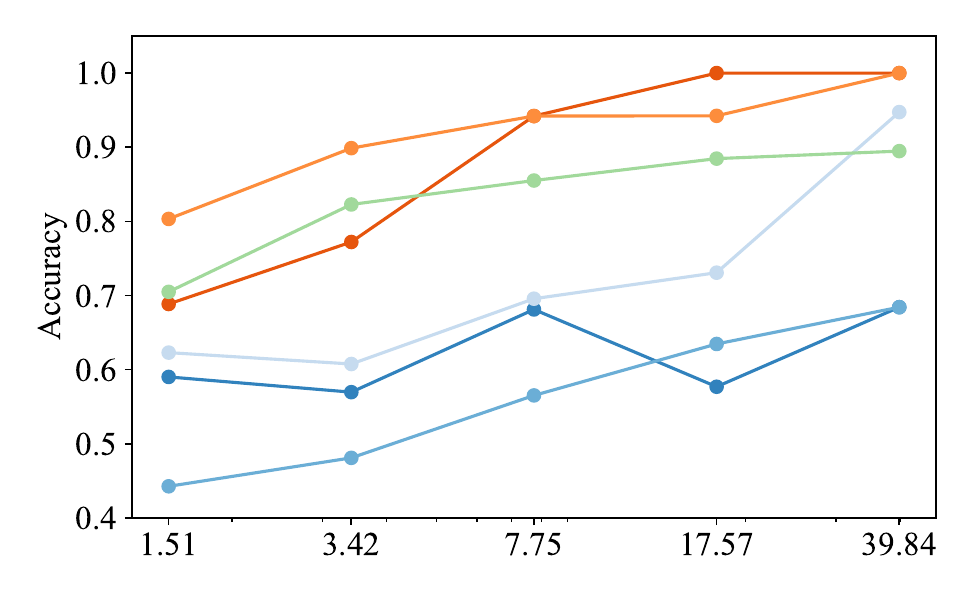}
        \caption{The model accuracy against the absolute difference in depth (meters) between the paired pixels.}
        \label{fig:depth_acc_distance}
    \end{subfigure}
    \begin{subfigure}[b]{0.225\linewidth}
        \centering
    \includegraphics[width=0.9\linewidth]{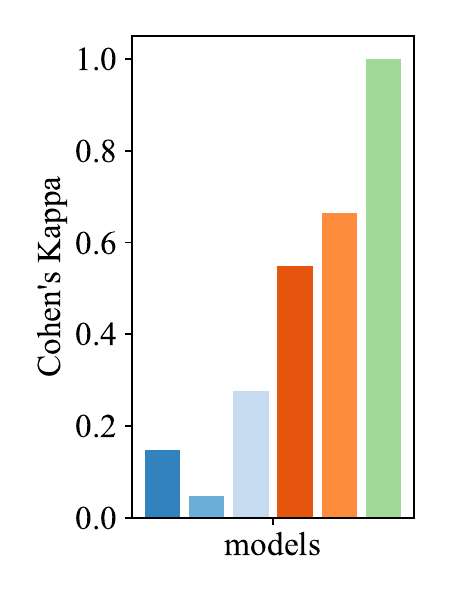}
        \caption{Cohen's $\kappa$ against humans.}
        \label{fig:depth_kappa}
    \end{subfigure}
    \hfill
    \begin{subfigure}[b]{0.76\linewidth}
        \centering
        \includegraphics[width=0.9\textwidth]{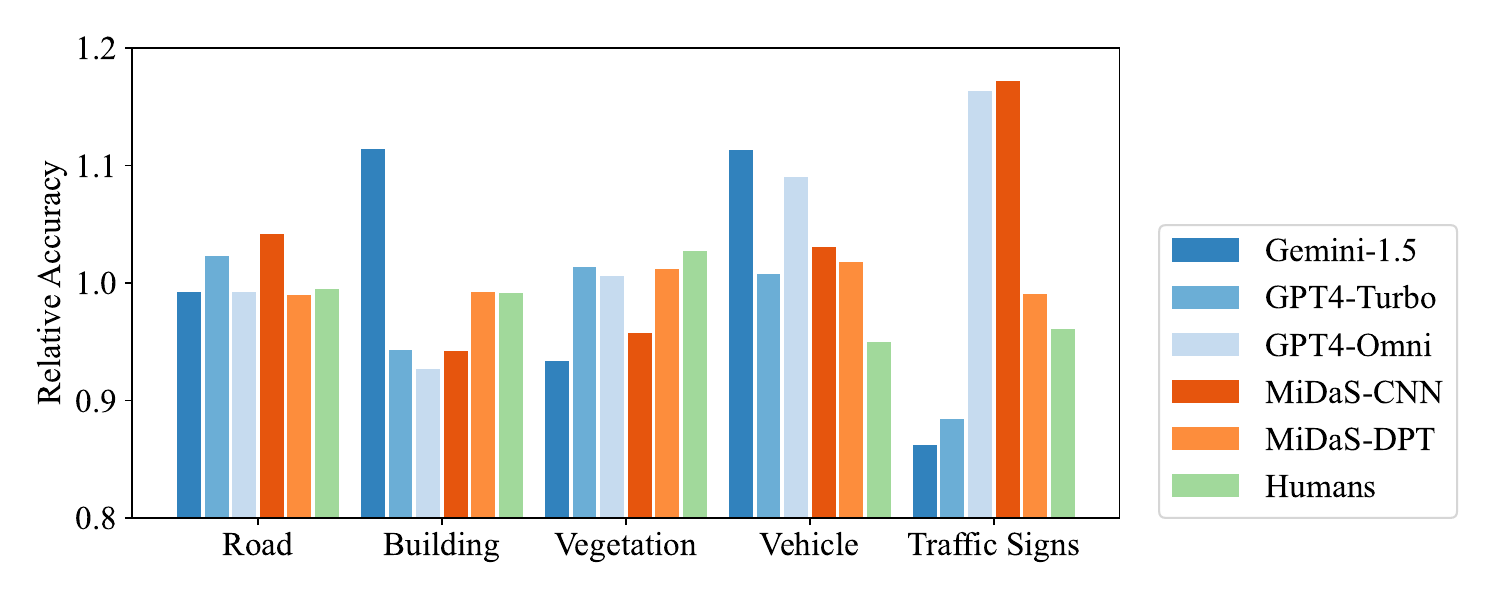}
        \caption{Accuracy against different semantic labels. The y-axis is normalized by the overall accuracy of each model.}
        \label{fig:depth_acc_classes}
    \end{subfigure}
    \hfill
    \vspace{-6mm}
    \caption{We compare the similarity between humans and different models using different metrics, including (a) pair sampling strategy, (b) relative depth difference, (c) Cohen's $\kappa$, and (d) semantic labels. Best viewed zoomed-in and in colors.}
    \label{fig:depth_similarity}
    \vspace{-7mm}
\end{figure}

\noindent\textbf{Model Accuracy} The primary results are shown in Fig.~\ref{fig:depth_acc_regular_flipped}. The two MiDaS models achieve the best performance, with the Transformer variant (MiDaS-DPT) being slightly better than the CNN variant. Surprisingly, both MiDaS variants outperform humans. This shows that state-of-the-art neural network models have competitive 3D understanding capabilities when trained on a large dataset, and can beat humans in unseen environments for 3D understanding tasks. 

All VLMs have significantly lower accuracy compared to MiDaS and humans, where the best-performing GPT4-Omni only achieves 67.9\% accuracy. We visualize the output of different methods in Fig.~\ref{fig:depth_acc_vis}. We find several failure modes in the VLM outputs: 1) \textit{localization}: markers placed on the road are classified as placed on a car; 2) \textit{scene understanding}: markers are understood as real objects in the scene and the depth is reasoned using relative size; and 3) \textit{reasoning}: the response is self-contradicting.

\noindent\textbf{Robustness} Results are shown in \cref{fig:depth_acc_regular_flipped}. The accuracy of all machine learning models drops significantly on the flipped images compared to regular ones (GPT4-Omni: 67.9\% $\rightarrow$ 53.5\%; MiDaS-DPT: 90.4\% $\rightarrow$ 73.9\%). In contrast, the human performance remains on par (82.1\% $\rightarrow$ 84.1\%) and is better than all models, showing the superior robustness of the human depth perception system. 

\noindent\textbf{Alignment with Humans} We measure the similarity of the answers of different models to the answers of humans, both qualitatively and quantitatively, as show in~\cref{fig:depth_similarity}.

Quantitative comparisons are shown in~\cref{fig:depth_kappa}. We use Cohen's $\kappa$~\cite{mchugh2012interrater} to compare the consistency between each model and humans. Note that Cohen's $\kappa$ rules out the effect of the model's accuracy, \textit{i.e.,} MiDaS won't have a higher alignment score just because it has higher overall accuracy. MiDaS-DPT achieves the best consistency, with Cohen's $\kappa$ of 0.66, followed by MiDaS-CNN of 0.56. All VLMs have very low consistency with humans. 

Qualitatively, we compare the distribution of the accuracy over different factors. In Fig.~\ref{fig:depth_acc_symmrand}, we compare the accuracy on symmetric and randomly sampled depth point pairs. The performance gap between the two cases is larger for the MiDaS-CNN compared to humans and MiDaS-DPT, showing the stronger alignment between Transformers and humans. Fig.~\ref{fig:depth_acc_distance} shows the accuracy against the difference in depth (meters) for the two-point sampled. The pairs with larger differences in depth are easier to tell apart, so slopes are positive. Comparing the shape of the curve, MiDaS-DPT is more similar to humans than MiDaS-CNN. Finally, in Fig.~\ref{fig:depth_acc_classes}, we compare the accuracy of each model on different classes. We merge the original KITTI classes into 5 super-classes. Note the accuracies are normalized by the overall accuracy of each model. The pattern of MiDaS-DPT is the most similar to humans, whereas MiDaS-CNN performs significantly better on traffic signs and worse on buildings and vegetation. In conclusion, MiDaS-DPT has the error pattern most similar to humans under multiple criteria, suggesting that Transformers are more similar to humans than CNNs in depth perception. \looseness=-1

\begin{figure*}[ht]
\vspace{-5mm}
    \centering
    \begin{subfigure}[b]{0.49\textwidth}
        \centering
         \includegraphics[width=0.8\textwidth]{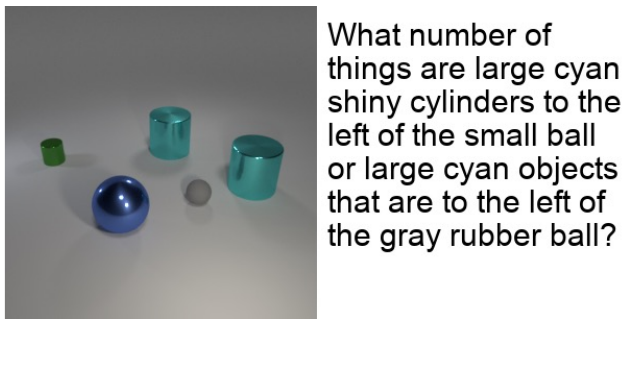}
        \caption{Example image and question pair.}
        \label{fig:clevr_example}
    \end{subfigure}
    \begin{subfigure}[b]{0.49\textwidth}
        \centering
         \includegraphics[width=0.8\textwidth]{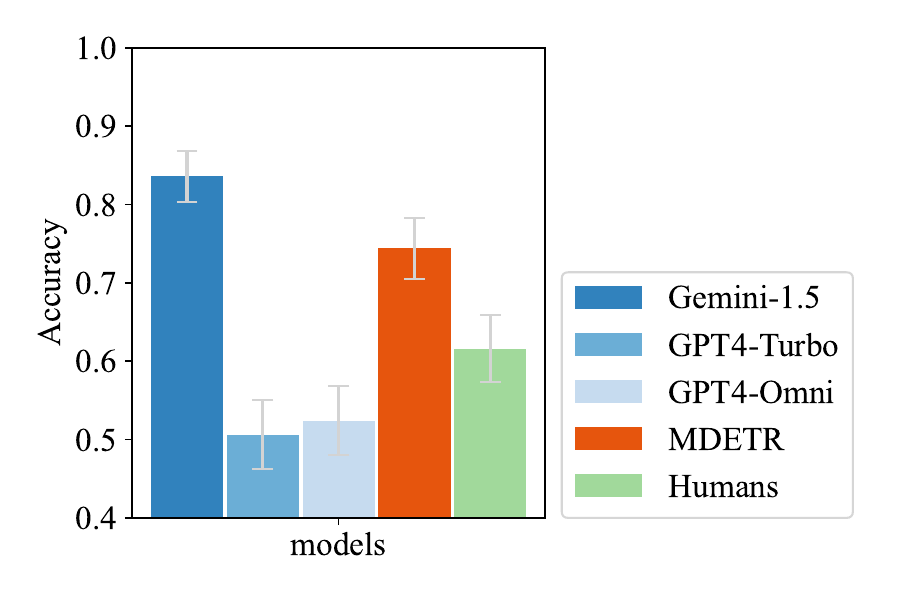}
        \caption{Accuracy on different models and humans.}
        \label{fig:clevr_acc}
    \end{subfigure}
    \hfill
    \begin{subfigure}[b]{0.49\textwidth}
        \centering
        \includegraphics[width=0.8\textwidth]{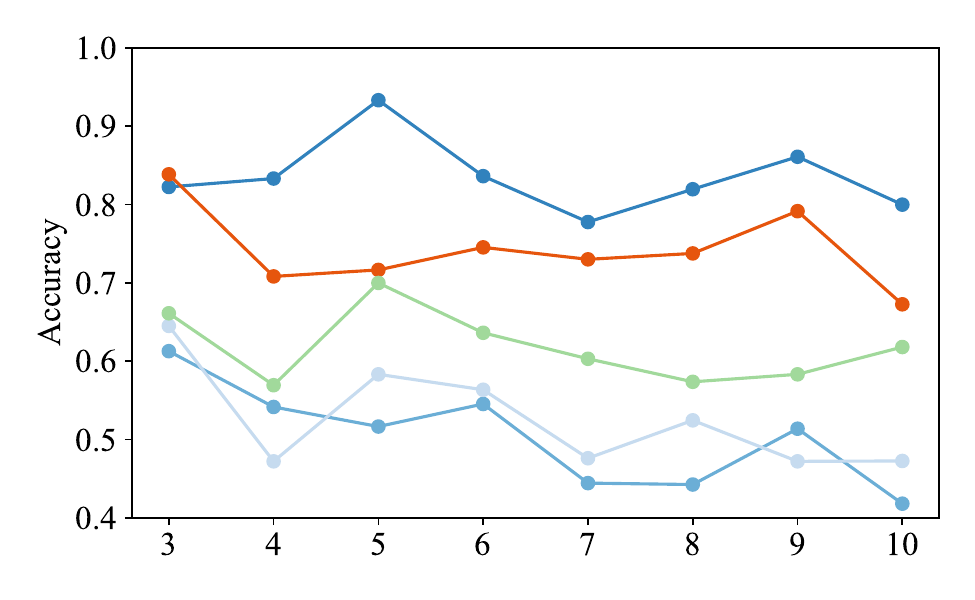}
        \caption{Accuracy against the number of objects in the scene.}
        \label{fig:clevr_numobj}
    \end{subfigure}
    \begin{subfigure}[b]{0.49\textwidth}
        \centering
    \includegraphics[width=0.8\textwidth]{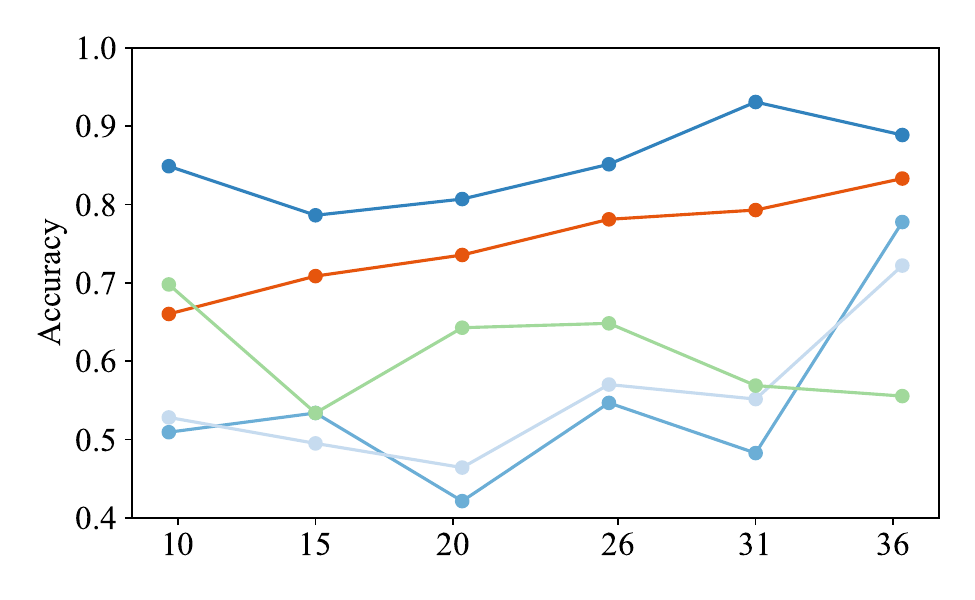}
        \caption{Accuracy against the length of the question.}
        \label{fig:clevr_sentencelength}
    \end{subfigure}
    \hfill
    \caption{Results on the spatial reasoning task. (a) Our benchmark requires a strong spatial reasoning ability and is very challenging. (b) Even the specialized VQA model MDETR can only achieve 74.4\% accuracy. (c) model accuracy drops as the scene complexity grows (more objects). (d) longer questions don't necessarily lead to worse performance. See text for detailed analysis.}
    \label{fig:clevr}
    \vspace{-4mm}
\end{figure*}

\begin{figure*}[ht]
    \centering
    \begin{subfigure}[t]{0.49\textwidth}
        \centering
    \includegraphics[width=\textwidth]{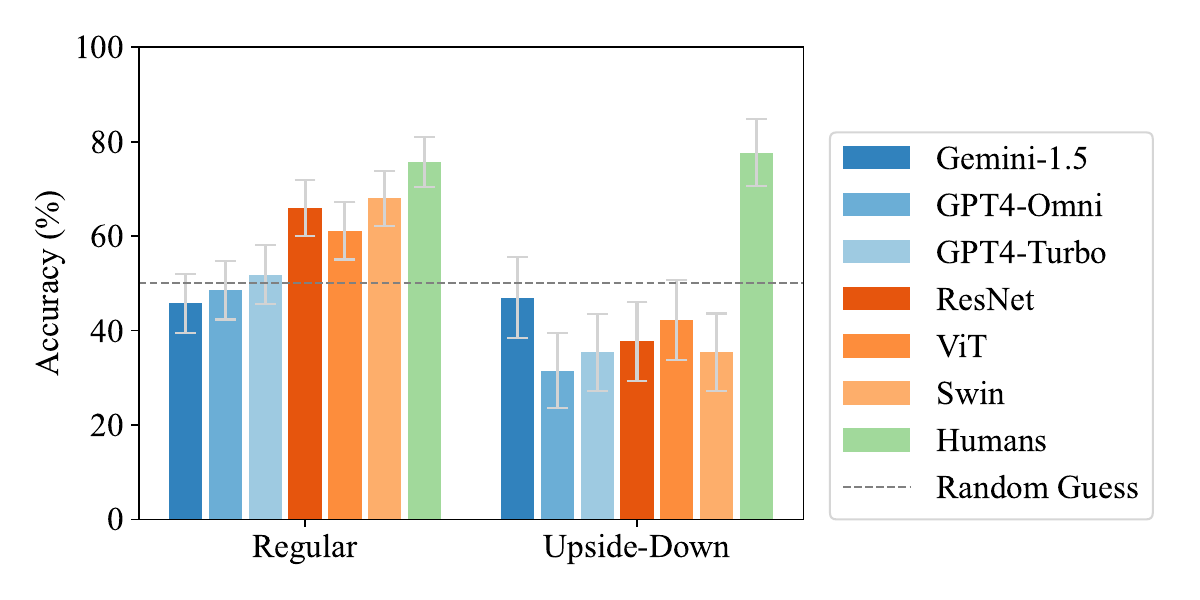}
        \caption{Relative camera pose classification accuracy.}
        \label{fig:cam_pose_acc}
    \end{subfigure}
    \hfill
    \begin{subfigure}[t]{0.49\textwidth}
        \centering
        \includegraphics[width=\textwidth]{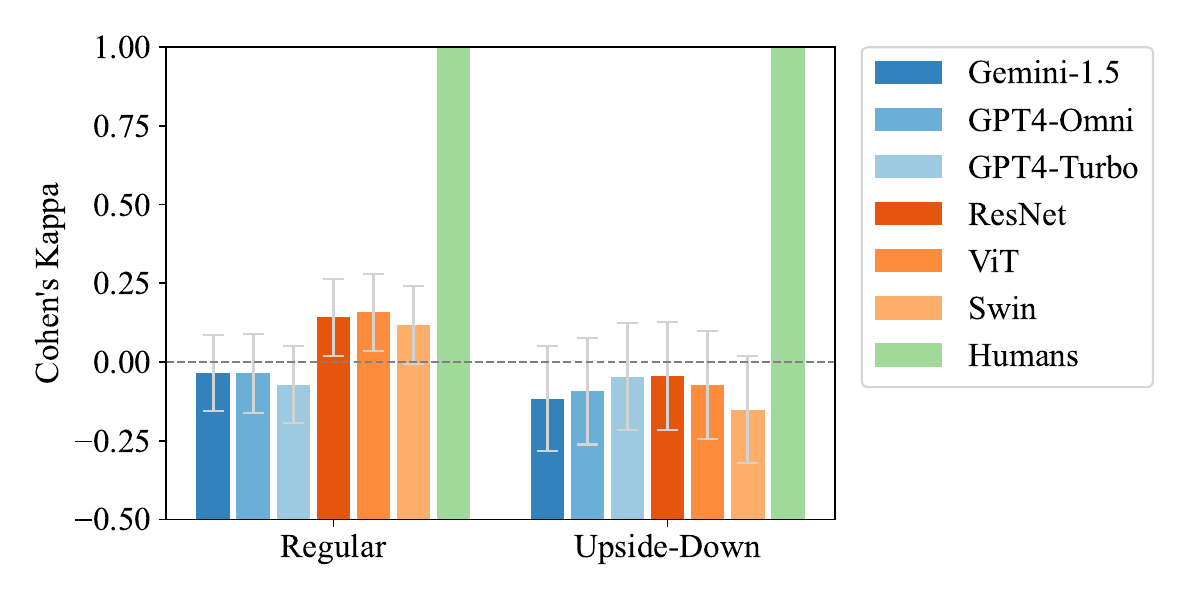}
        \caption{Cohen's $\kappa$ similarity to human relative pose classification.}
        \label{fig:cam_pose_sim}
    \end{subfigure}
    \hfill
    \caption{Comparison between specialist neural networks, LVMs, and humans on relative camera pose classification. The bars are 95\% confidence intervals.}
    \label{fig:camera_pose}
    \vspace{-2mm}
\end{figure*}

\subsection{Spatial Reasoning}

\textbf{Models Compared} In addition to the three VLMs and humans, we evaluate against MDETR~\cite{kamath2021mdetr} as the specialized model on our benchmark. MDETR is a state-of-the-art VQA model trained on CLEVR~\cite{johnson2017clevr}.

\noindent\textbf{Model Accuracy} The overall accuracy of different models and humans are shown in \cref{fig:clevr_acc}. Human accuracy is 61.6\%. The accuracy is low for two reasons: 1) Our benchmark is challenging, requiring reasoning through a long logical chain. See \cref{fig:clevr_example} for an example. 2) While we believe humans can do better if they pay full attention and are given enough time, the numbers we report are the ``average'' humans on MTurk, instead of the performance upper bound. 

Among the VLMs, Gemini-1.5 achieves an accuracy of 83.6\%, which is surprisingly even better than MDETR (74.4\%) which is trained on the CLEVR training set. Although the 3D understanding capability of VLMs may be limited as shown in the other tasks, their reasoning capability seems to be very strong, achieving good performance on the reasoning-oriented VQA task. The two GPT4 models perform relatively poorly, with GPT4-Omni having a slightly better accuracy of 52.4\%.

\noindent\textbf{Alignment with Humans} We compare the accuracy of different models against the scene complexity, measured by the number of objects in the scene. Results are shown in \cref{fig:clevr_numobj}. The accuracy of all models drops as the scene complexity increases, while human accuracy is not much affected. We also measure the complexity of the question. While it is difficult to define the exact complexity, we use the number of words in a sentence as a proxy. The results shown in \cref{fig:clevr_sentencelength} are quite counter-intuitive: the accuracy of all models grows as the question becomes more complex. In contrast, human accuracy drops slightly. None of the models show a strong correlation with humans, highlighting the potentially different ways that models and humans approach the spatial reasoning task.

\begin{figure*}[!ht]
\vspace{-3mm}
    \centering
        \begin{subfigure}[t]{0.32\linewidth}
        \centering
        \includegraphics[width=\linewidth]{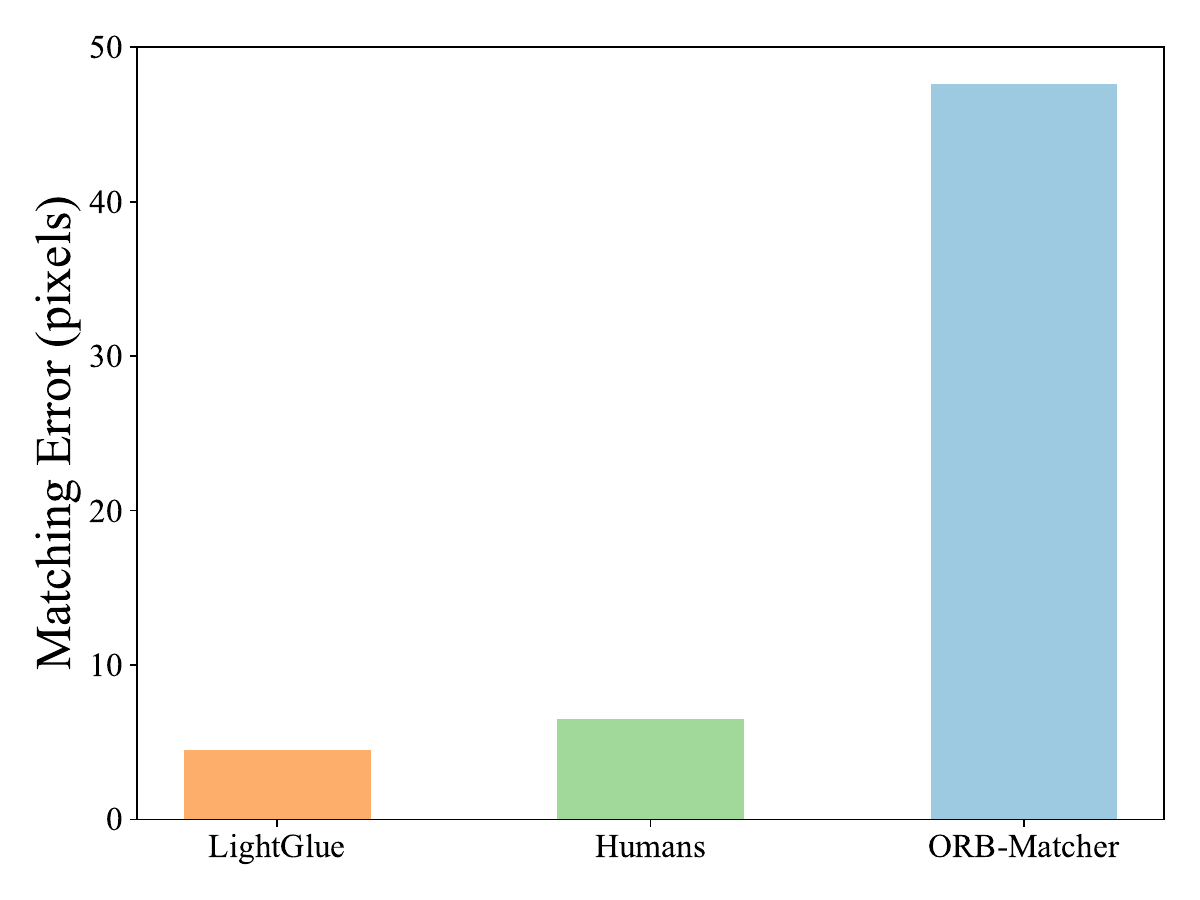}
        \caption{Matching error made by humans, LightGlue, and ORB matcher measured EPE.}
        \label{fig:matching/acc}
    \end{subfigure}
    \hfill
        \begin{subfigure}[t]{0.32\linewidth}
        \centering
        \includegraphics[width=\textwidth]{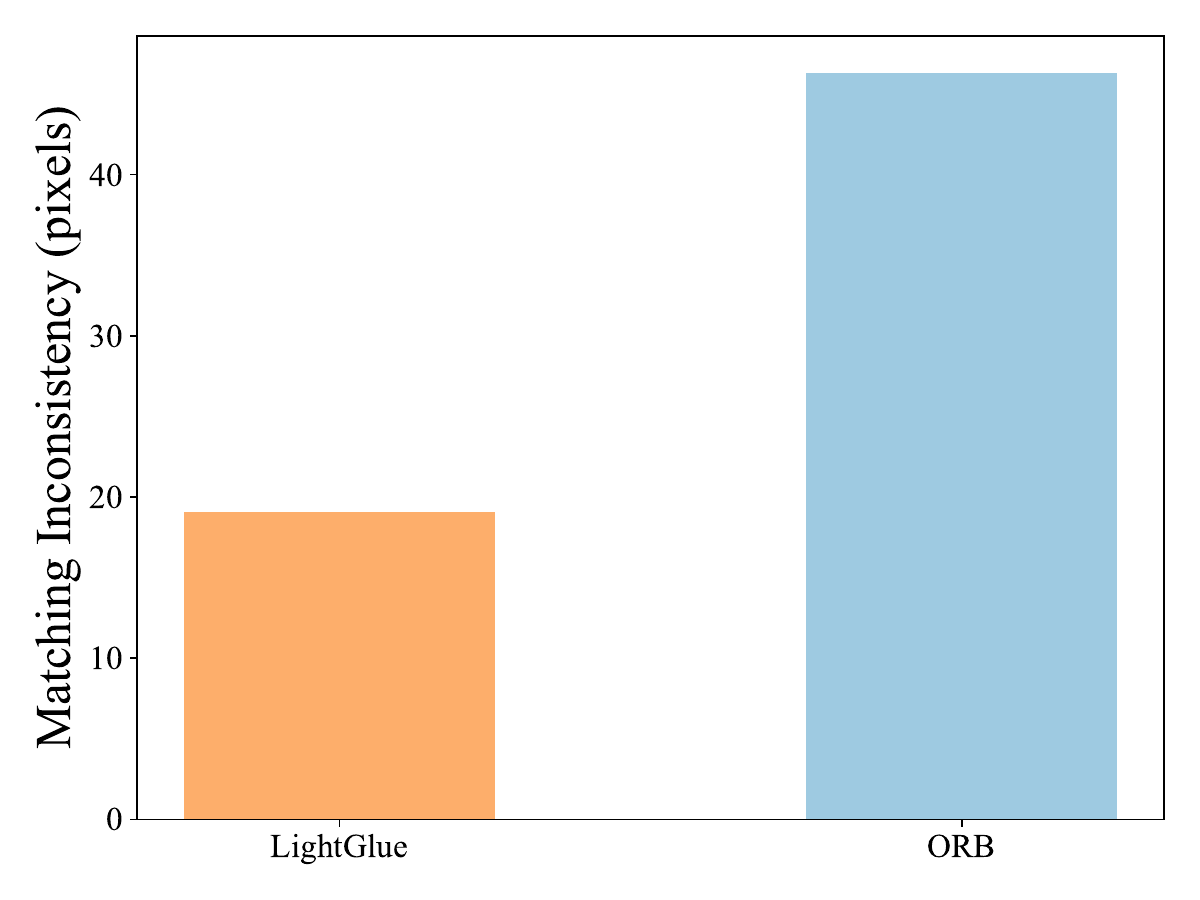}
        \caption{Matching inconsistency score with humans. LightGlue is more similar to human matching.}
        \label{fig:matching/similarity}
    \end{subfigure}
        \begin{subfigure}[t]{0.32\textwidth}
        \centering
        \includegraphics[width=\linewidth]{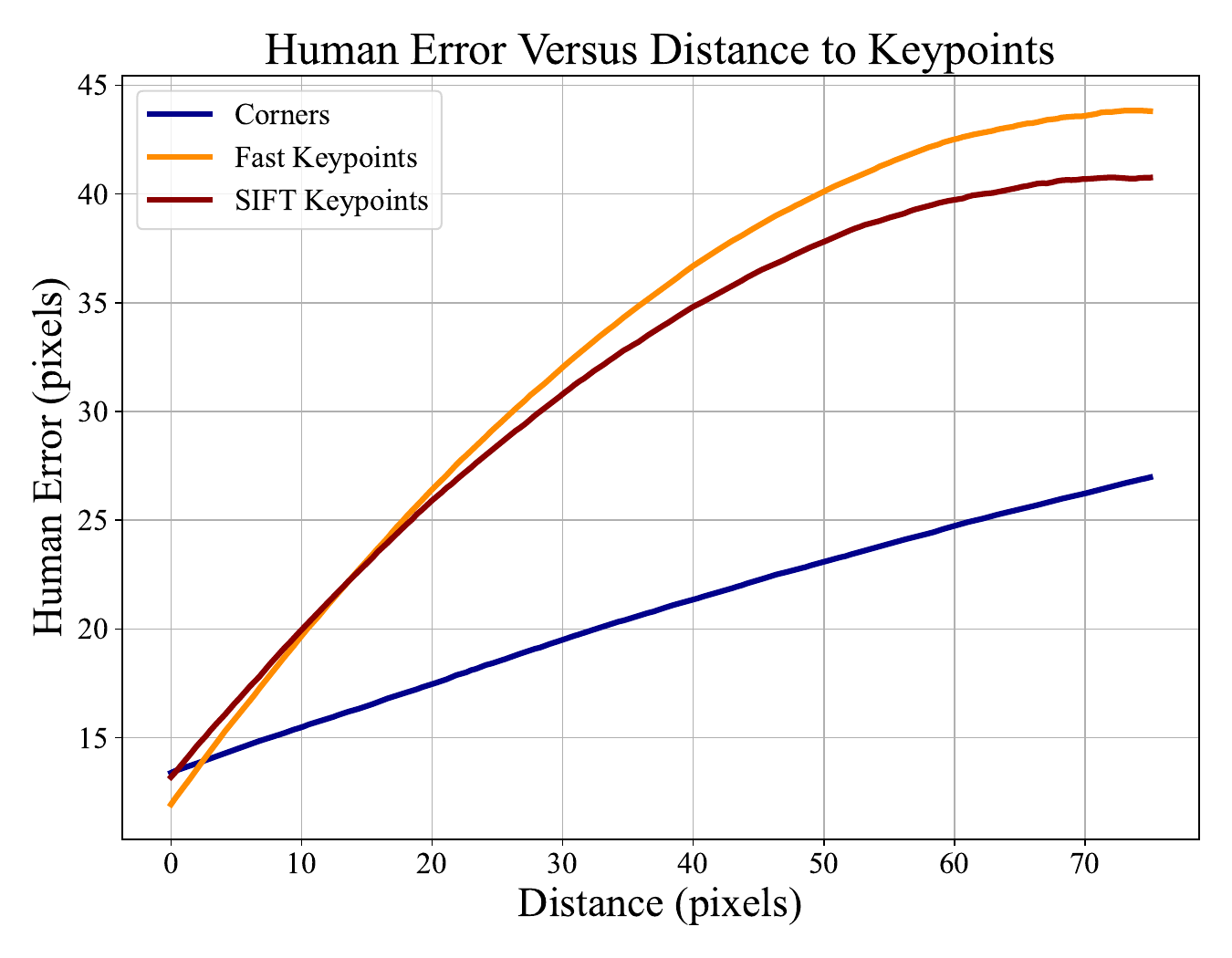}
        \caption{Human error and distance to detector keypoints.}
        \label{fig:matching/distance}
    \end{subfigure}
    \hfill
    \caption{Matching experiment results. Transformer-based LightGlue is more similar to human matching than the classical ORB matcher.}
    \label{fig:matching}
    \vspace{-3mm}
\end{figure*}

\begin{figure}[!ht]
    \centering\includegraphics[width=0.9\linewidth]{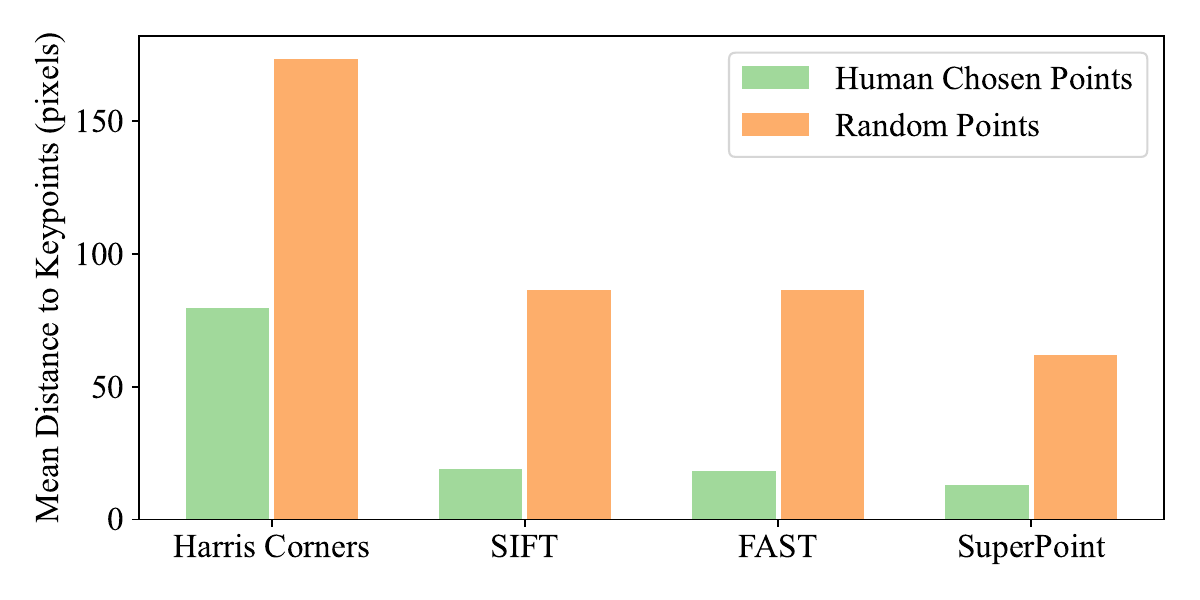}
    \vspace{-3mm}
        \caption{The average distance of human chosen vs random keypoint to the nearest detector keypoint. Humans are more likely to choose salient points than randomly guess. }
        \label{fig:keypoints_error_all}
        \vspace{-5mm}
\end{figure}

\subsection{Relative Camera Pose Estimation}
\textbf{Models Compared} We train our custom specialized models on the BlendedMVS~\cite{yao2020blendedmvs} dataset. We use ResNet-50~\cite{he_deep_2015}, ViT~\cite{dosovitskiy_image_2021}, and Swin Transformer~\cite{liu_swin_2021} with pretrained ImageNet~\cite{deng_imagenet_2009} weights as our neural network backbone for comparison. We provide the most prominent movement axis (up/down vs. right/left) to the subject as input and formulate the question as a two-way classification task. We ask the same question to both human subjects and VLMs.

\noindent\textbf{Model Accuracy} The main results are shown in \cref{fig:camera_pose}. Humans achieve the highest accuracy (75.7\%) followed by the specialized model with a Swin Transformer backbone (68\%). Interestingly, the VLM models perform approximately the same as random guess, the most accurate one being GPT4-Turbo with 51.8\% accuracy. 

\noindent\textbf{Robustness} Although specialized models have similar accuracy to humans with regular images, they perform significantly worse than random guess when the images are flipped (ResNet: 65.9\% $\rightarrow$ 37.7\%; ViT: 61.1\% $\rightarrow$ 42.3\%; Swin: 68\% $\rightarrow$ 35.4\%). This observation holds for VLMs as well (GPT4-Omni: 48.6\% $\rightarrow$ 31.5\%; GPT4-Turbo: 51.8\% $\rightarrow$ 35.4\%), with the exception of Gemini-1.5 (45.7\% $\rightarrow$ 46.9\%) which has similar accuracy. None of the VLMs perform better than random guess when the images are flipped. In contrast, the human accuracy is almost invariant to geometric perturbation (75.7\% $\rightarrow$ 77.7\%). This suggests that the human visual system is way more robust compared to both VLMs and specialized models in the context of camera pose estimation.

\noindent\textbf{Alignment with Humans} We measure Cohen's $\kappa$ to compare the models in terms of how similar they are to humans in the answers they output. \cref{fig:cam_pose_sim} shows that ViT is slightly more similar to humans than CNNs in this task, although the 95\% confidence intervals highly overlap. VLMs have negative Cohen's $\kappa$ with humans, showing that their answers are highly dissimilar even controlling for accuracy. When the images are flipped, all models have negative Cohen's $\kappa$, which demonstrates that all models are dissimilar to humans in terms of output under geometric perturbation. In conclusion, none of the models significantly align with humans, suggesting that they are unsuitable as models of human perception in this task.

\subsection{Keypoint Matching}

\textbf{Models Compared} The keypoint matching task is further divided into two sub-tasks, \textit{i.e.,} keypoint detection and correspondence matching. They are usually performed by different models in the keypoint matching literature~\cite{detone_superpoint_2018, sarlin2020superglue}. 
\begin{itemize}[leftmargin=*]
    \item \textbf{Keypoint Detection} We compare against a set of classical hand-cratfed detectors, including Harris Corner Detector~\cite{Harris1988ACC} with the top $k$ corners, Difference of Gaussians (DoG) used in SIFT descriptors~\cite{sift}, and the FAST detectors~\cite{Rosten2006MachineLF} used in ORB descriptors~\cite{Rublee2011ORBAE}. We also compare the neural network based detector SuperPoint~\cite{detone_superpoint_2018}. All detectors are configured to choose the same number of best candidates for a fair comparison. 
    \item \textbf{Matching} We use LightGlue~\cite{lindenberger_LightGlue_2023} for our deep learning based method of end-to-end matching and ORB-matcher for our classical vision based method. LightGlue uses DISK~\cite{tyszkiewicz_disk_2020} for feature description, and uses Transformer as its backbone.
\end{itemize}

\noindent\textbf{Model Accuracy} We evaluate human matches, LightGlue, and ORB matching on the average end-point error (EPE) from the ground truth correspondence. \cref{fig:matching/acc} shows that LightGlue makes the least matching errors, followed closely by humans, while the ORB matcher is much less accurate. This suggests that the state-of-the-art model can achieve comparable performance to humans on the matching task, which requires detailed localization capability.

\noindent\textbf{Alignment with Humans}
\begin{itemize}[leftmargin=*]
    \item \textbf{Keypoint Detection} We investigate what kinds of points humans are more likely to choose when asked to find matching points between two images. We find that humans have a tendency towards choosing salient points. \cref{fig:keypoints_error_all} shows that the subjects were much more likely to choose corners, SIFT keypoints, FAST keypoints, or SuperPoint keypoints than random choice. Among the detectors, SuperPoint has the best alignment with human keypoint selection. This shows that neural networks have higher similarity to humans than hand-crafted detectors.

    \item \textbf{Matching} We evaluate the consistency of LightGlue and ORB matcher with human matches. We measure consistency by computing the average difference in EPE made by humans and models on each keypoint. As shown in \cref{fig:matching/similarity}, LightGlue (diff=19.0px) is more consistent with human matches compared to ORB (diff=44.3px). This suggests that human 3D visual system is more similar to a Transformer-based neural network architecture that performs global reasoning compared to a classical algorithm that keeps track of local statistical information. 

    Finally, We analyze what kind of keypoints are difficult for humans to match accurately. We study the relationships between human errors and the distance between the keypoints they choose to the closest keypoints detected by models. \cref{fig:matching/distance} shows that as the keypoints selected by humans stray away from the salient points, the matching error gets larger regardless of whether saliency is measured by corners, FAST, or SIFT. This finding is consistent with the common belief that pixels with rich textures such as corners are easier to match.
    
    Linear regression analysis shows that around 15\% of the variance in human error is explained by the distance from SIFT and FAST keypoints with $p < 0.001$. We also find that approximately 10\% of the variance is explained by the distance from a corner with $p < 0.05$. 
    
\end{itemize}

\section{Limitations}
There are a few limitations in our analysis: 1) While we focus on the four tasks that we consider most fundamental for 3D vision—relative depth, spatial reasoning, relative camera pose estimation, and keypoint matching—it would be beneficial to evaluate additional tasks, such as surface normals. 2) Due to time and resource constraints, we only evaluated closed-source VLMs. Evaluating open-source VLMs could provide a more comprehensive comparison. 

\section*{Acknowledgements}
This work was partially supported by the National Science Foundation (2331763). We thank Thomas A. Langlois for insightful discussions.

{
    \small
    \bibliographystyle{plain}
    \bibliography{main}
}

\clearpage

\setcounter{section}{0}
\renewcommand{\thesection}{\Alph{section}}

\begin{flushleft}
\section*{Appendix}
\end{flushleft}

\section{Human Annotation Collection}

\begin{figure*}[ht!]
    \centering
    \begin{subfigure}[t]{0.49\textwidth}
        \centering
    \includegraphics[width=\textwidth]{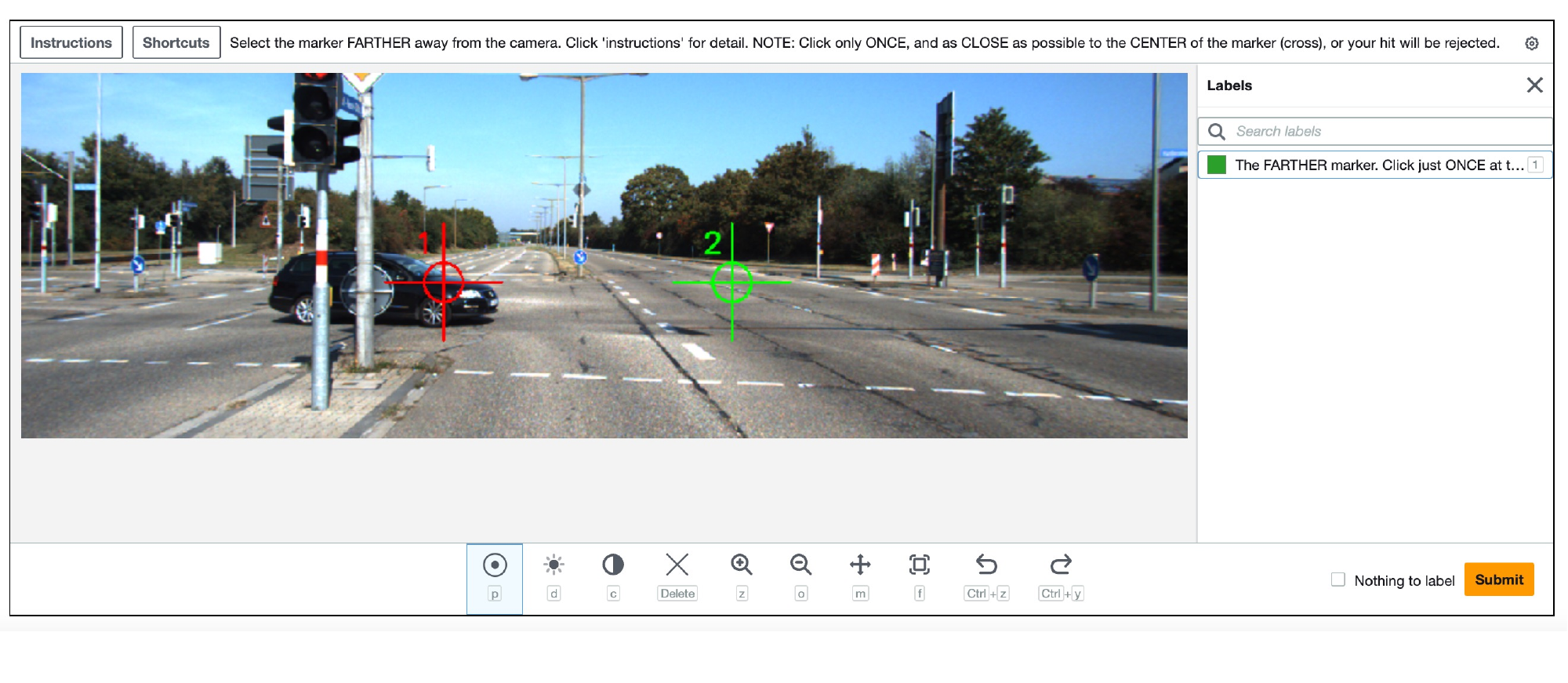}
        \caption{Annotation Interface. The user is asked to click near the center of the maker that they think is farther away from the camera. The user can use the MTurk's built-in functionalities to zoom and move if necessary.}
        \label{fig:depth_mturk_interface}
    \end{subfigure}
    \hfill
    \begin{subfigure}[t]{0.49\textwidth}
        \centering
        \includegraphics[width=\textwidth]{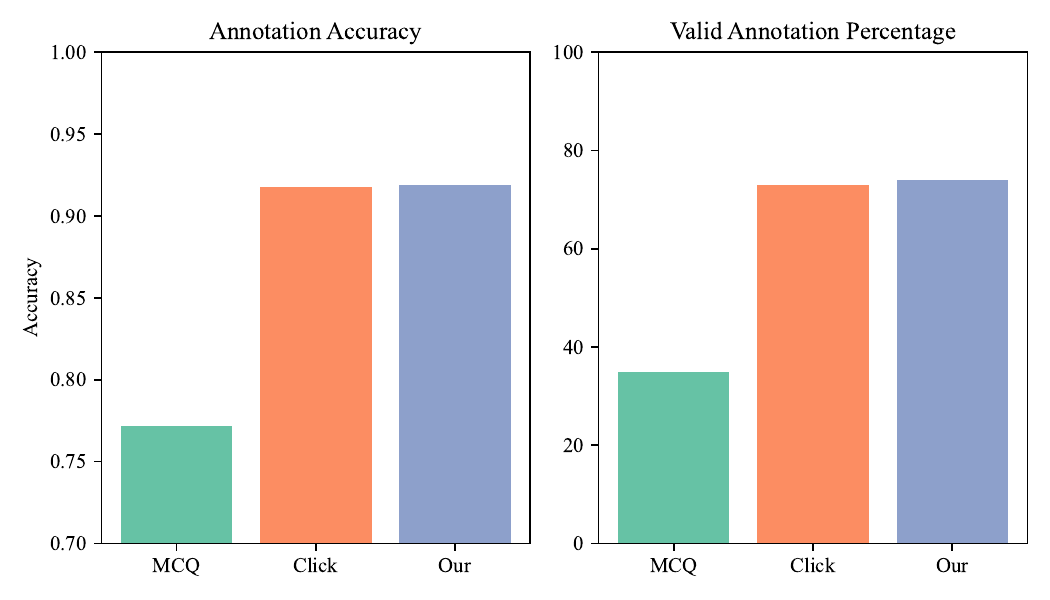}
        \caption{Multiple-choice questions (MCQ) cannot detect the bots, resulting in only 35\% valid data and 77\% accuracy. In comparison, using the click-based interface boosts the valid percentage and accuracy to 73\% and 91\% respectively, being very close to the human upper-bound (annotate ourselves).}
        \label{fig:depth_mturk_quality}
    \end{subfigure}
    \hfill
    \begin{subfigure}[t]{0.49\textwidth}
        \centering
        \includegraphics[width=\textwidth]{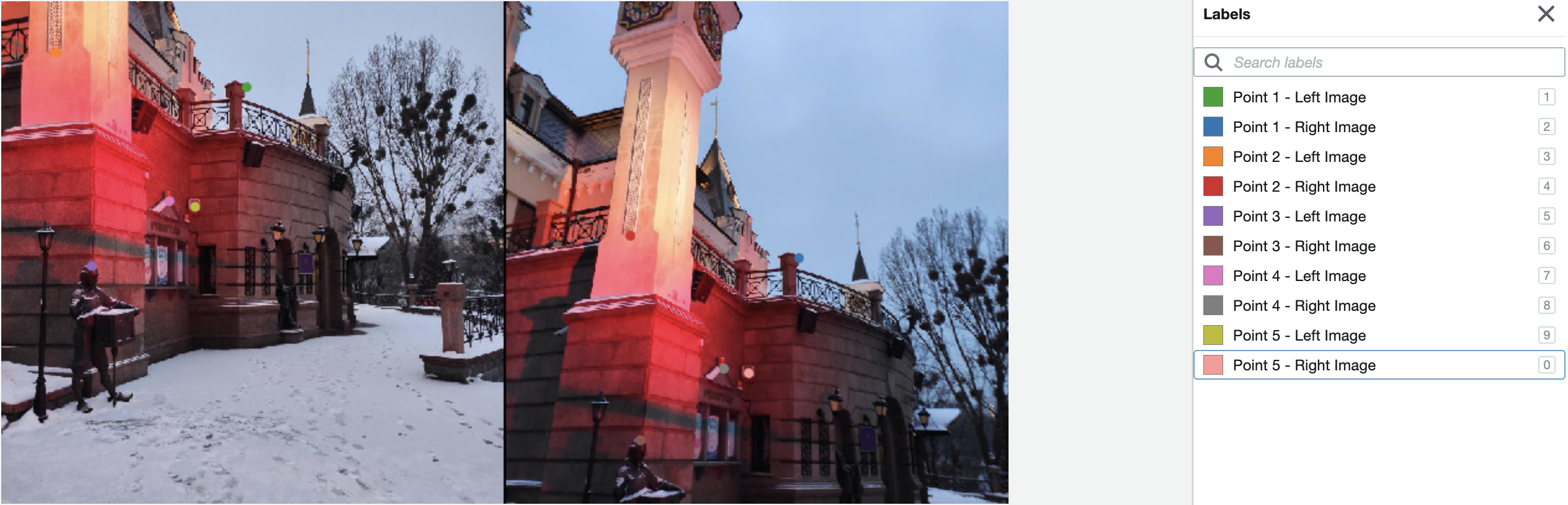}
        \caption{The user interface we use to collect keypoint detection and matching annotations from human subjects.}
        \label{fig:keypoint_ui}
    \end{subfigure}
    \hfill
    \begin{subfigure}[t]{0.49\textwidth}
        \centering
        \includegraphics[width=\textwidth]{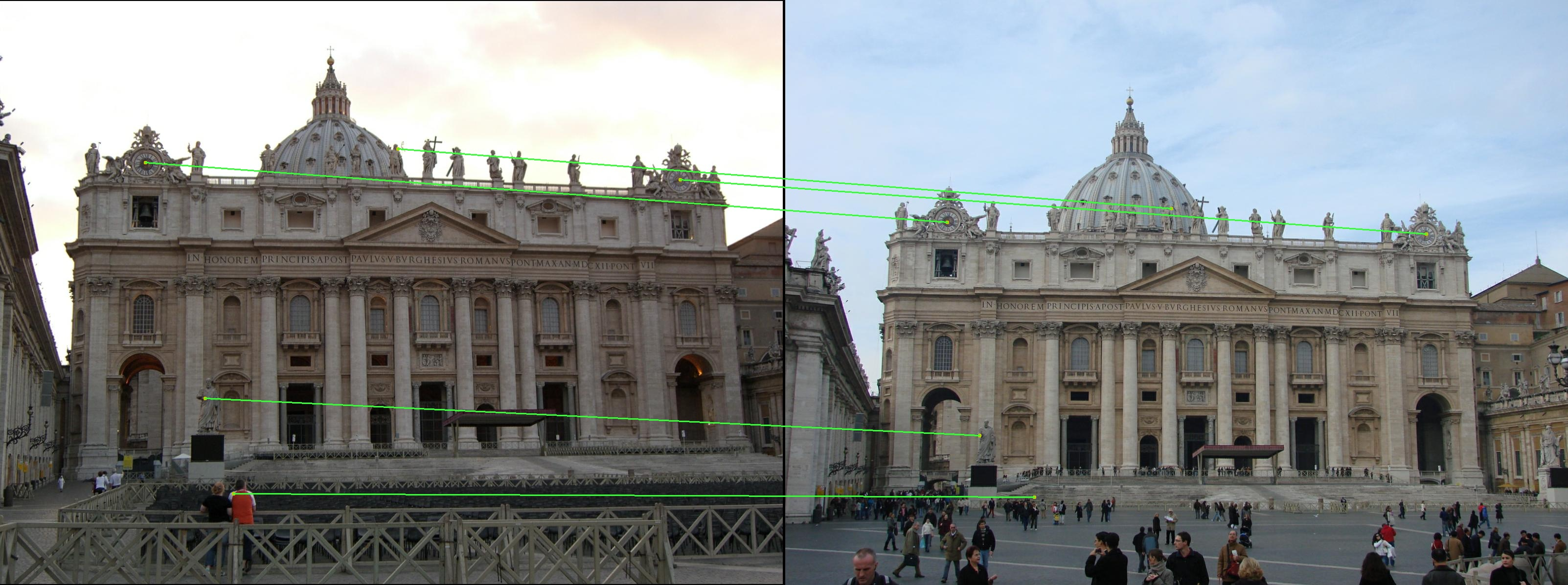}
        \caption{Example matches collected from human annotators.}
        \label{fig:correspondence}
    \end{subfigure}
        \hfill
    \begin{subfigure}[t]{0.49\textwidth}
        \centering
        \includegraphics[width=\textwidth]{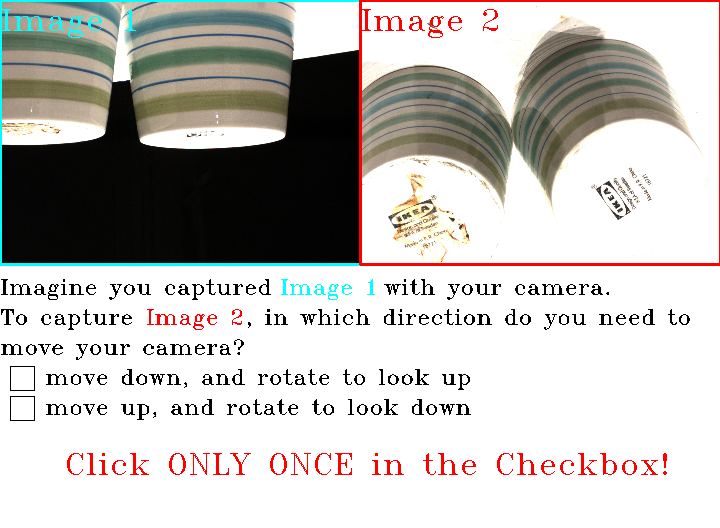}
        \caption{Camera pose estimation user interface. This is an example of a flipped image. }
        \label{fig:Q_visualization}
    \end{subfigure}
    \caption{Human annotation interface on MTurk and annotation quality control.}
    \label{fig:depth_mturk}
\end{figure*}

We provide more details of the human annotation interface on MTurk, as shown in \cref{fig:depth_mturk}.

In \cref{fig:depth_mturk_interface} we show the interface that the workers see on MTurk. Instead of completing a multiple-choice question, the workers are asked to click near the center of the marker that they think is closer to the camera. This allows us to effectively filter out the bots/spammers on the MTurk platform. Our quality control is effective as shown in \cref{fig:depth_mturk_quality}. With the clicking questions, we are able to boost the human annotation accuracy from 77\% to 91\%, which is almost the same accuracy as what we get by annotating ourselves.

\begin{figure*}[t]
    \includegraphics[width=\linewidth]{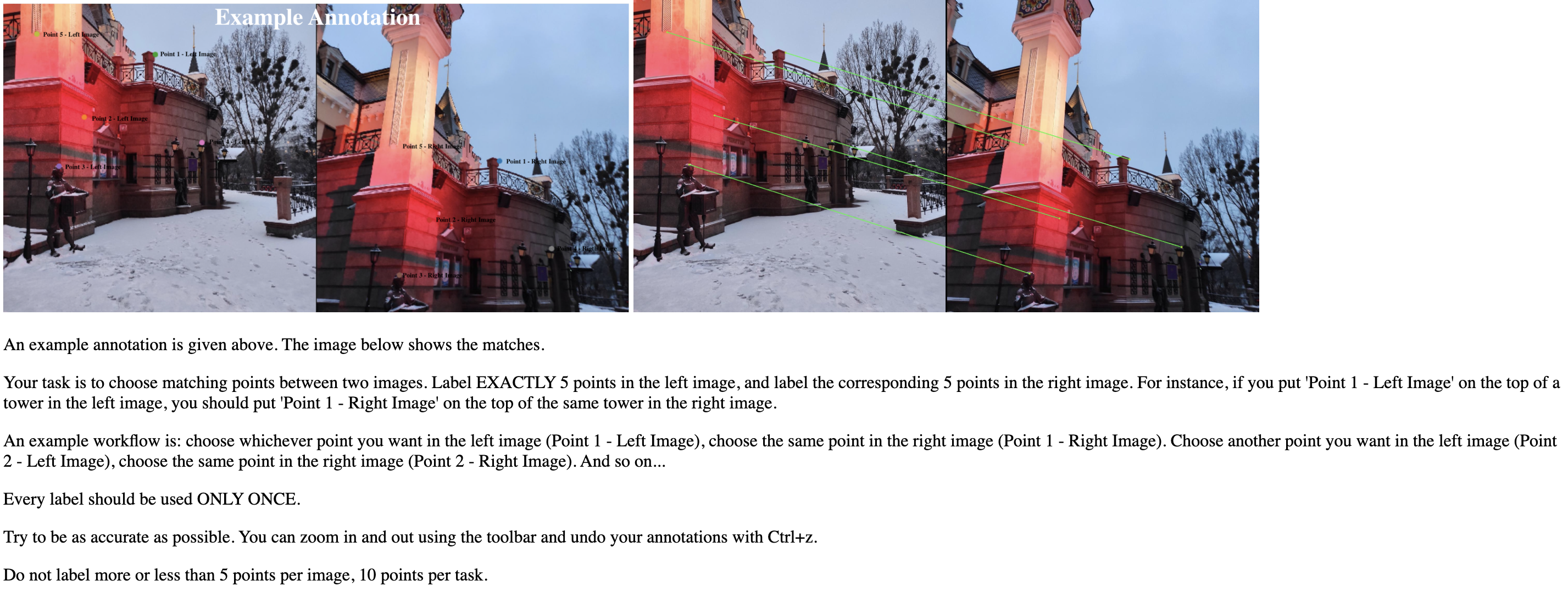}
    \caption{Full instruction prompt given to human subjects for keypoint detection and matching annotations. }
    \label{fig:keypoint_instructions}
\end{figure*}

For the keypoint detection and matching task, we prompt the workers with a pair of co-visible images from the Megadepth-1500 \cite{sun_loftr_2021, li_megadepth_2018} dataset. Our short prompt is ``Choose EXACTLY 5 points in the left image and the same 5 points in the right image." See Fig. \ref{fig:keypoint_instructions} for the full instructions given to the subjects. Also, see Fig. \ref{fig:keypoint_ui} for the user interface we use to collect annotations. 

\cref{fig:correspondence} shows an example correspondence annotation we collected for the keypoint matching task. To ensure quality, we only keep responses where the subject followed all instructions. For instance, we eliminate responses where the subject did not label exactly 5 keypoints in each image. We also eliminate responses where the subject matched a keypoint to another keypoint in the same image or to a completely random point in the other image.

\begin{figure*}[!ht]
    \centering
    \begin{subfigure}[b]{0.32\textwidth}
        \centering
    \includegraphics[width=\textwidth]{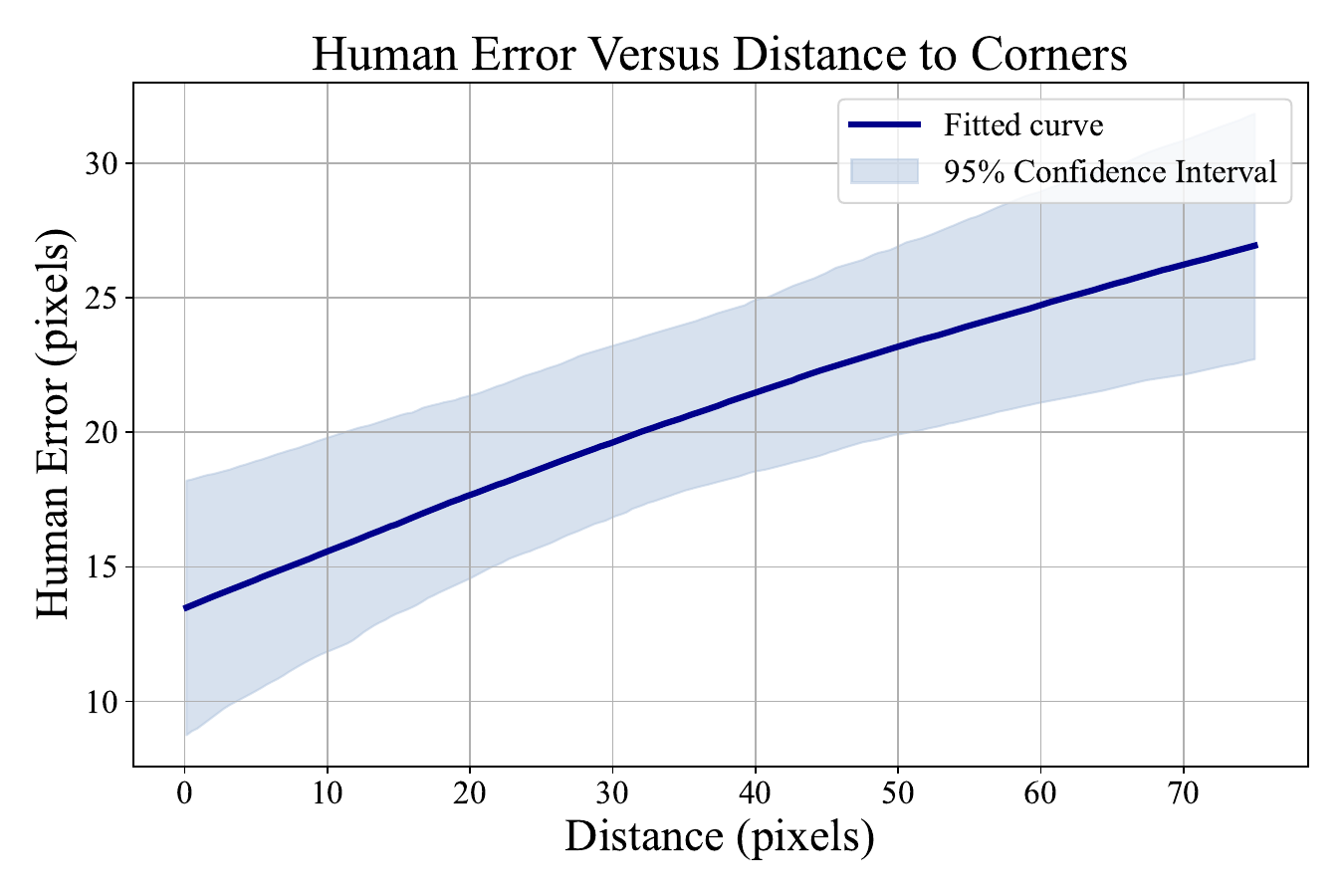}
        \caption{Human error and distance to corners.}
        \label{fig:error_corner}
    \end{subfigure}
    \hfill
    \begin{subfigure}[b]{0.32\textwidth}
        \centering
        \includegraphics[width=\textwidth]{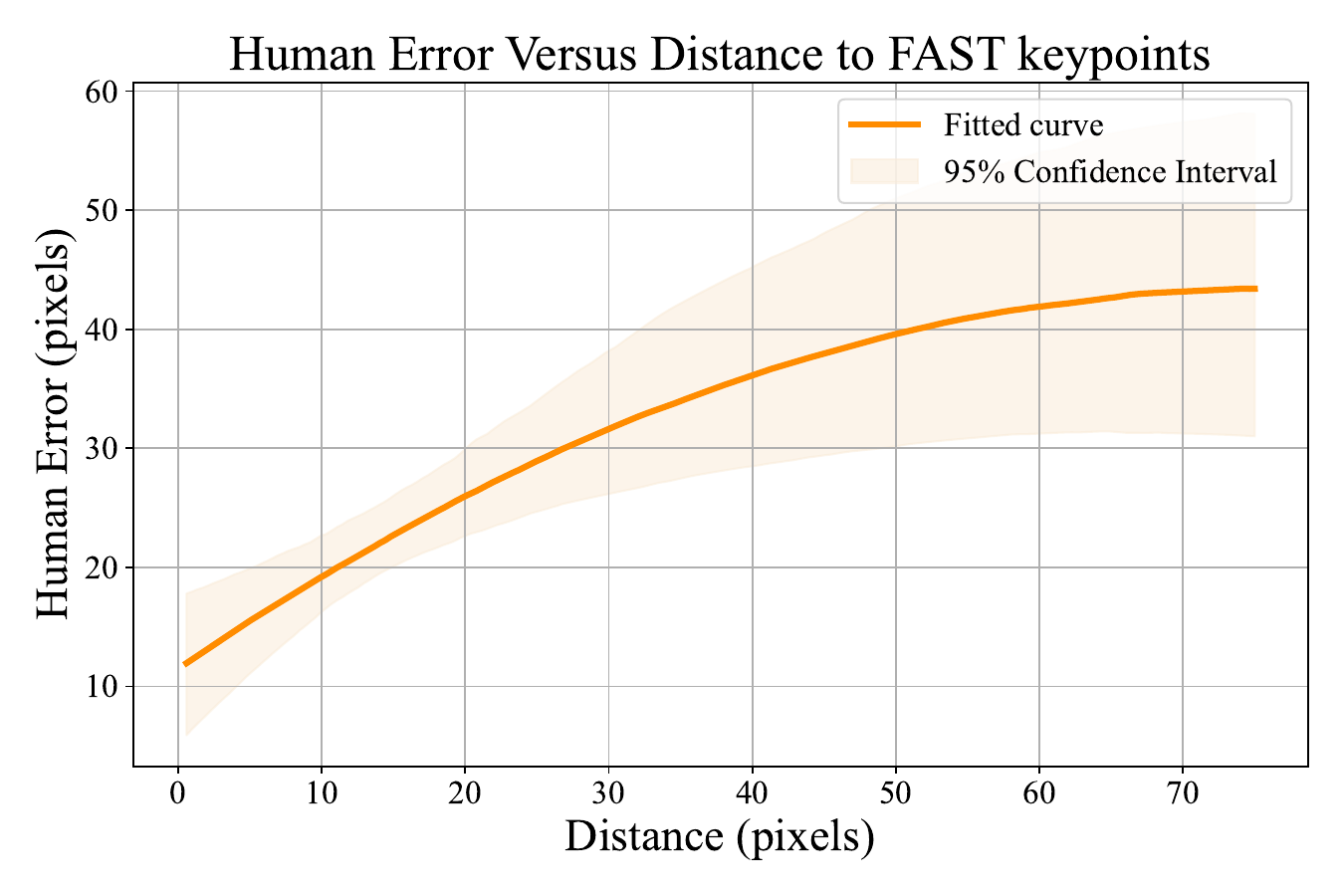}
        \caption{Human error and distance to FAST keypoints.}
        \label{fig:error_fast}
    \end{subfigure}
    \hfill
    \begin{subfigure}[b]{0.32\textwidth}
        \centering
        \includegraphics[width=\textwidth]{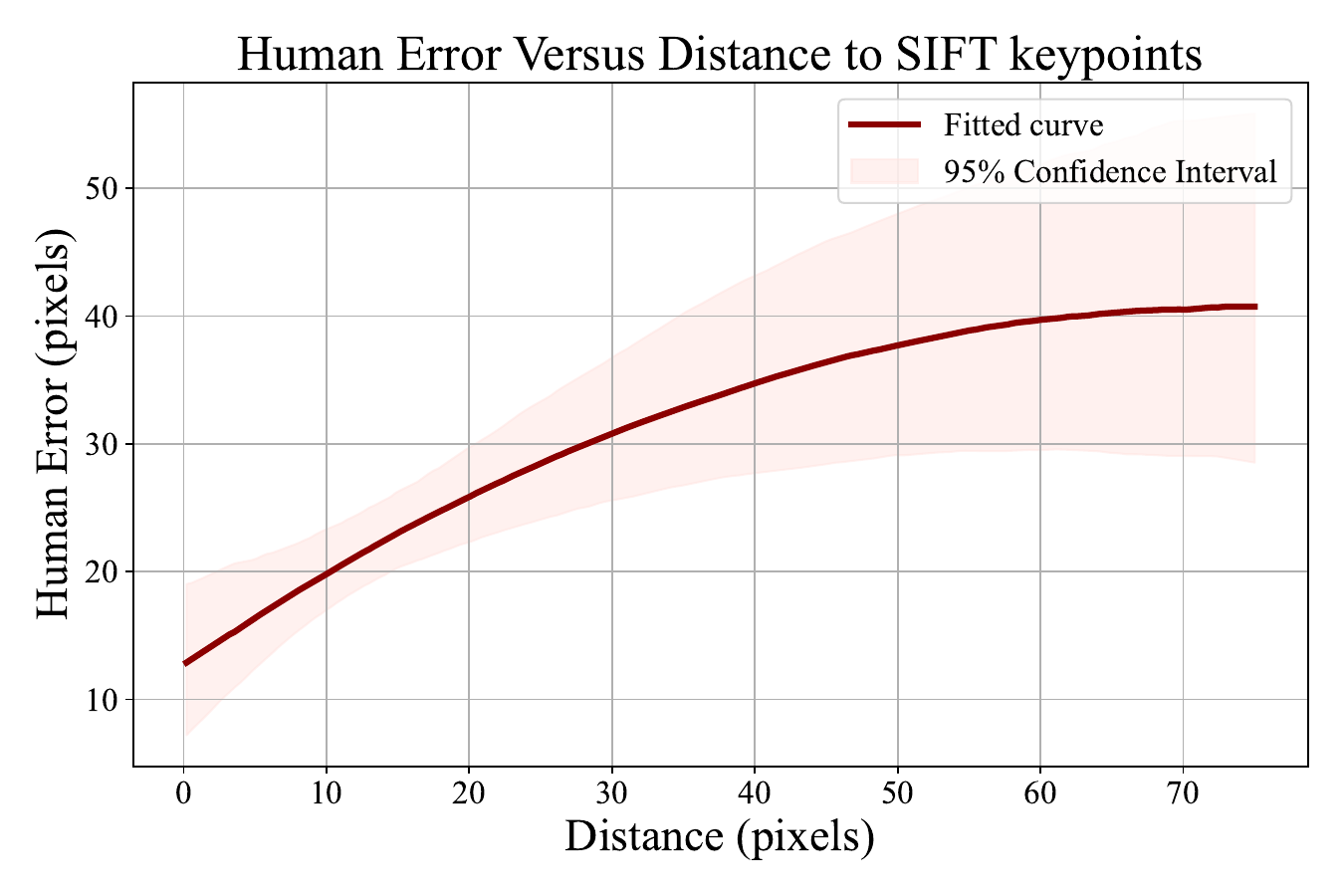}
        \caption{Human error and distance to SIFT keypoints.}
        \label{fig:error_sift}
    \end{subfigure}

        \caption{Matching errors humans make with respect to the ground truth correspondence. Subjects make fewer mistakes when they match points that are salient. (a) The closer a point is to a corner, the easier it is to match for humans. (b) The closer a point is to a FAST keypoint, the easier it is to match for humans. (c) The closer a point is to a SIFT keypoint, the easier it is to match for humans.  }
        \label{fig:human_error_individual}
\end{figure*}

\begin{figure*}[t]

        \begin{subfigure}[b]{0.49\textwidth}
        \centering
        \includegraphics[width=\linewidth]{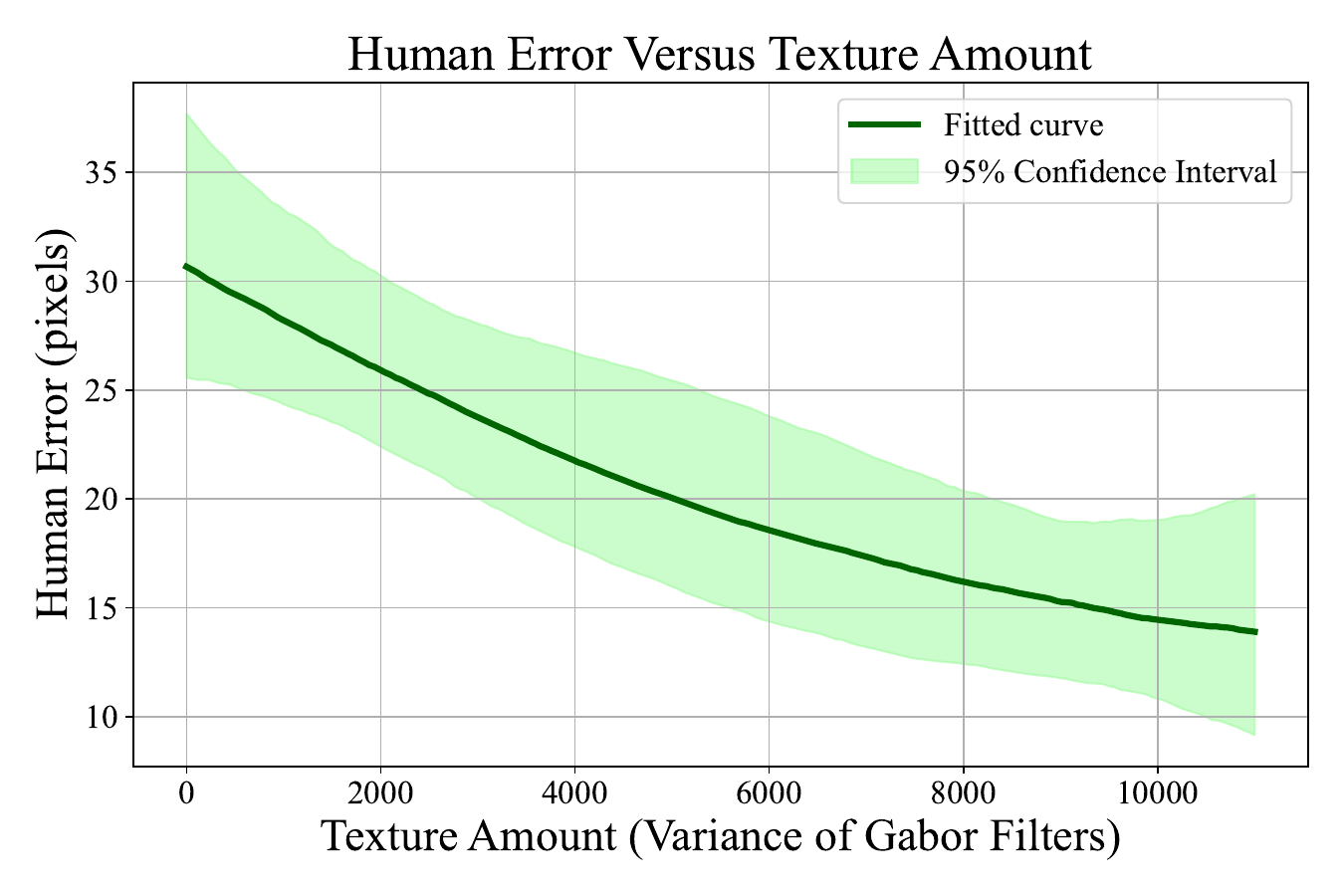}
        \caption{Human error and the amount of texture.}
        \label{fig:error_texture}
    \end{subfigure}
    \hfill
    \begin{subfigure}[b]{0.49\textwidth}
        \centering
        \includegraphics[width=\textwidth]{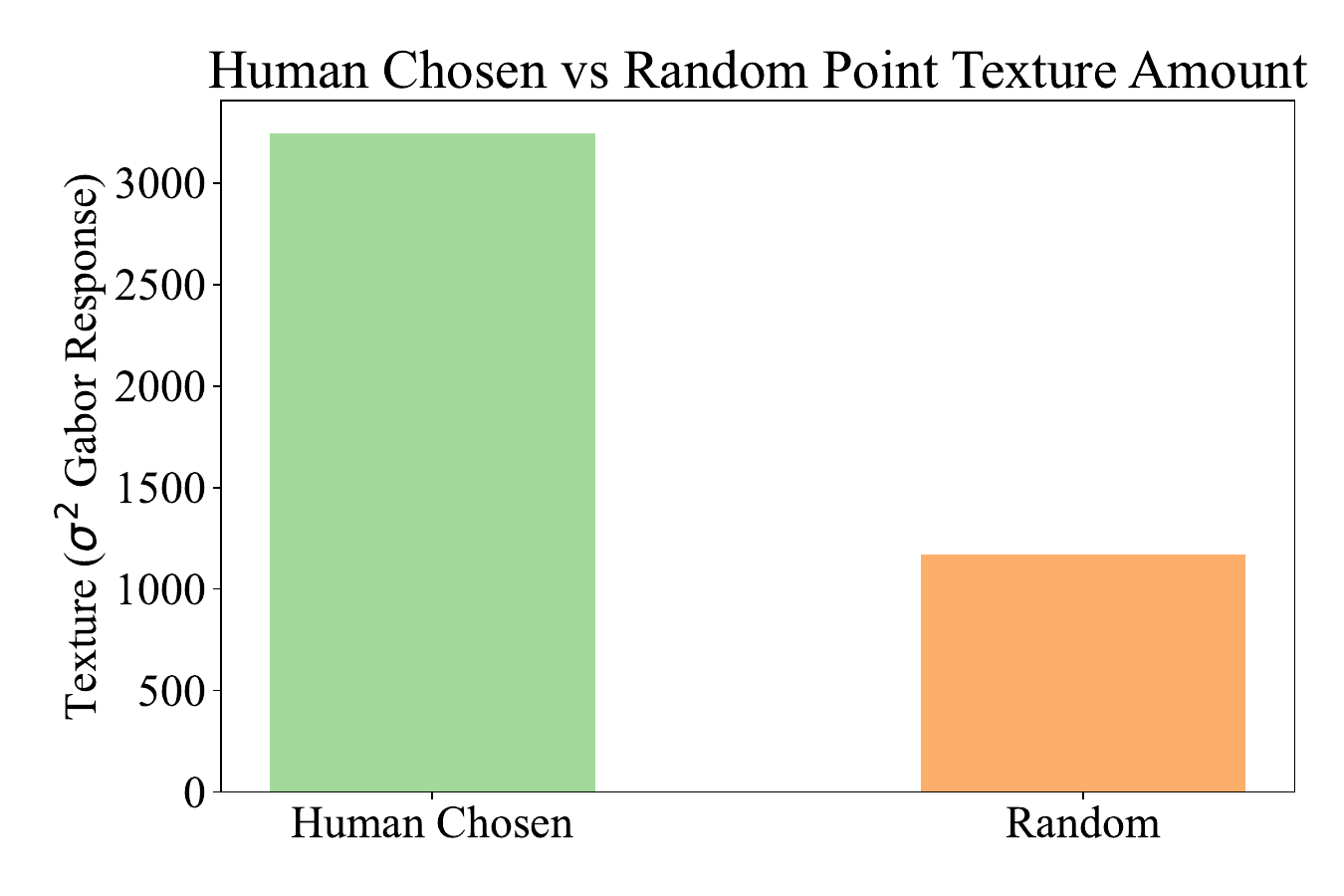}
        \caption{The average amount of texture around keypoints chosen by the subjects versus the amount of texture around a random point. }
    \label{fig:texture_preference}
    
    \end{subfigure}
    \caption{(a) More textured locations are easier to match for humans. (b) The subjects are more likely to choose textured keypoints}
    \label{fig:human_error_all}
\end{figure*}

\section{More Results on Keypoint Matching}

We analyze what causes human subjects to make errors in keypoint matching, which could hopefully inform benchmarks and training datasets that rely on human annotations. Fig. \ref{fig:human_error_individual} shows that subjects were more accurate around keypoints that are detected by SIFT \cite{sift}, FAST \cite{Rosten2006MachineLF}, and Harris Corner detectors. In other words, if a subject matched a keypoint that is close to a corner, for instance, they would be closer to the ground-truth correspondence. We further observe in Fig. \ref{fig:error_fast} and Fig. \ref{fig:error_sift} that the end-point error (EPE) increases logarithmically with the distance to the nearest SIFT and FAST keypoints. However, as the confidence intervals in Fig. \ref{fig:human_error_individual} reveal, these conclusions are not as clear for very distant keypoints due to the lack of samples. Our linear regression analysis shows that around 15\% of the variance in human error is explained by the distance from SIFT and FAST keypoints with $p < 0.001$. We also find that approximately 10\% of the variance is explained by the distance from a corner with $p < 0.05$. 

We analyze how the human annotations are affected by texture as well. We observe that human subjects tend to overwhelmingly choose textured points compared to random choice. Fig. \ref{fig:texture_preference} demonstrates this phenomenon. To measure how textured the patch around a pixel is, we use a combination of Gabor filters \cite{feichtinger_gabor_1998} with different orientations. We take the variance of these filters to measure the amount of texture around a pixel. We further observe in Fig. \ref{fig:error_texture} that the subjects made less matching errors when matching textured points, meaning that human EPE was lower for textured keypoints.

\section{Experimental Setup: Camera Pose Estimation}
To train the neural network models, we set up the camera pose estimation as a two-way classification problem, with the most prominent axis of movement as input. This is because the most prominent axis is given to VLMs and Humans as input and they are asked to classify the movement direction, so we mimic the same setup in order to ensure fair evaluation. Given a pair of images, we convert the ground-truth relative pose between them into the ground-truth primary move direction. Specifically, given the x,y components of the relative translation vector between the two frames $\mathbf{T} = [T_x, T_y]$, we first compute the absolute values of the components $\mathbf{A} = [|T_x|, |T_y|]$. We then identify the component with the largest magnitude by $\text{index} = \text{argmax}(\mathbf{A})$, which indicates the axis along which the most significant movement occurs. Based on the sign of this component, the ground truth answer is given as follows:
\[
D = \begin{cases} 
   0 & \text{if } \text{index} = 0 \text{ and } T_x > 0 \quad (\text{+x direction})\\
   1 & \text{if } \text{index} = 0 \text{ and } T_x < 0 \quad (\text{-x direction})\\
   0 & \text{if } \text{index} = 1 \text{ and } T_y > 0 \quad (\text{+y direction})\\
   1 & \text{if } \text{index} = 1 \text{ and } T_y < 0 \quad (\text{-y direction})
\end{cases}
\]

We train Resnet, ViT, and Swin Transformer backbones on this classification task. Given a pair of images, we pass each of them through the backbone and concatenate the two feature vectors. We concatenate this feature vector with the index given above which encodes the primary movement axis. We then pass the concatenated feature vector through an MLP output head two predict the primary movement direction. We use cross-entropy loss to train each network. As our training dataset, we choose the BlendedMVS \cite{yao2020blendedmvs} dataset and train each network for 15 epochs on NVIDIA RTX 3090 GPUs. 

During testing, we evaluate both the networks and VLMs on a two-way classification task where the objective is to distinguish between the direction of the movement along the primary movement axis. If the primary movement is along the x-axis, we ask VLMs ``Imagine you captured image 1 with your camera. To capture image 2, in what direction do you need to move your camera? A: move left, and rotate to your right; B: move right, and rotate to your left?". If the primary movement is along the y-axis, we ask VLMs ``Imagine you captured image 1 with your camera. To capture image 2, in what direction do you need to move your camera? A: move down, and rotate to look up; B: move up, and rotate to look down". Note that these are the same questions asked to the human subjects. For the test dataset, we use DTU \cite{jensen2014large}. We deliberately choose the test-stage to be zero-shot for the neural networks by not fine-tuning them on DTU. The goal here is to compare humans, VLMs, and neural networks in equal conditions assuming none of the VLMs have been trained on DTU.

\end{document}